\documentclass{article}

\usepackage{arxiv}
\usepackage[utf8]{inputenc} 
\usepackage[T1]{fontenc}    
\usepackage{hyperref}       
\usepackage{url}            
\usepackage{booktabs}       
\usepackage{amsfonts}       
\usepackage{nicefrac}       
\usepackage{microtype}      
\usepackage{lipsum}		

\usepackage{graphicx}     
\usepackage{subcaption}
\usepackage{amsmath}

\title{Statistical Agnostic Mapping: a Framework in Neuroimaging based on Concentration Inequalities}

\author{
  J.M. Gorriz\thanks{gorriz@ugr.es, jg528@cam.ac.uk} \\
  Department of Signal Theory and Communications\\
  University of Granada\\
  Granada, Spain \\
   \And
 SiPBA Group \\
  Department of Signal Theory and Communications\\
  University of Granada\\
  Granada, Spain\\
  \texttt{sipba@ugr.es} \\
     \And
 Cam Neuroscience Group \\
  Department of Psychiatry\\
  University of Cambridge\\
  Cambridge, UK\\
  \texttt{jg825@cam.ac.uk} \\
       \And
       International Initiatives\\
for the Alzheimer's Disease Neuroimaging Initiative (ADNI)\\
for the Parkinson’s Progression Markers Initiative (PPMI)\\
}

\begin{document}
\maketitle

\begin{abstract}
In the 70s a novel branch of statistics emerged focusing its effort in selecting a function in the pattern recognition problem, which fulfils a definite relationship between the quality of the approximation and its complexity. These data-driven approaches are mainly devoted to problems of estimating dependencies with limited sample sizes and comprise all the empirical out-of sample generalization approaches, e.g. cross validation (CV) approaches. Although the latter are \emph{not designed for testing competing hypothesis or comparing different models} in neuroimaging, there are a number of theoretical developments within this theory which could be employed to derive a Statistical Agnostic (non-parametric) Mapping (SAM) at voxel or multi-voxel level. Moreover, SAMs could relieve i) the problem of instability in limited sample sizes when estimating the actual risk via the CV approaches, e.g. large error bars, and provide ii) an alternative way of Family-wise-error (FWE) corrected p-value maps in inferential statistics for hypothesis testing. In this sense, we propose a novel framework in neuroimaging based on concentration inequalities, which results in (i) a rigorous development for model validation with a small sample/dimension ratio, and (ii) a less-conservative procedure than FWE p-value correction, to determine the brain significance maps from the inferences made using small upper bounds of the actual risk.
\end{abstract}


\section{Introduction}
In the last decades neuroscience has transitioned from qualitative case reports to quantitative, longitudinal and multivariate population studies in the quest for defining the abnormal patterns of disease pathogenesis. Neuroscience has recently provided valuable insight by means of classical statistics, e.g  statistical inference based on null-hypothesis ($H_0$) testing or regression-type analyses. Thus, brain mapping community has predominantly used null hypothesis testing for exploratory analyses in whole brain searches \cite{Friston2013}. In this context, classical inference makes emphasis on in-sample image-based statistical estimates from previously assumed data models to determine the existence of relevant effects (large or subtle) in binary group comparisons. The critical p-value (significance vs. not significant) is often \emph{complemented by effect-size measures} of the magnitude of the phenomenon $H_0$ \cite{Bzdok17}.

On the other hand, out-of sample generalization approaches in machine learning (ML), such as Cross-Validation (CV), try to estimate on unseen new data the actual error of the classifier in the (binary) classification problem. Despite the methods and goals of predictive CV inference are distinct from classical extrapolation procedure \cite{Lindquist13}, they are actually exploited within statistical frameworks aimed at providing statistical significance \cite{Reiss15}. Examples are bootstrapping, binomial or permutation (``resampling'') tests, etc. which have demonstrated to be competitive outside the comfort zone of classical statistics, filling \emph{otherwise-unmet inferential needs}. 

In the pattern classification problem we usually assume the existence of classes ($H_1$) that are differentiated by the classifiers in terms of accuracy (Acc) on a presumably independent dataset. Empirical confidence intervals or plausible Acc values derived from CV are consequently used to evaluate the system performance and to conclude (improperly in a statistical sense) $H_1$. Moreover, in limited sample sizes the most popular K-fold CV method \cite{Kohavi95} has demonstrated to sub-optimally work under unstable conditions \cite{Gorriz18,Gorriz19,Varoquaux18} and then, the predictive power of the fitted classifiers can be arguable.

Beyond the latter techniques, CV in ML is well-framed into a data-driven statistical learning theory (SLT) which is mainly devoted to problems of estimating dependencies with limited amounts of data \cite{Vapnik82}. Although, CV-ML approaches were not originally designed to test hypothesis in brain mapping \cite{Friston2013}, they are theoretically grounded to provide maps of \emph{confidence intervals} (protected inference). As shown in present study, this can be achieved by assessing the upper bounds of the actual error in a binary classification problem, and by using simple significance tests for a population proportion. Thus, assessing with high probability the quality of the fitting function (and its generalization ability) in terms of in/out-sample predictions can be conceptualised, under a hypothesis testing scenario, as the inverse problem of ``carefully rejecting $H_0$'', that is, the problem of rejecting $H_1$ and thus accepting $H_0$ (there is no effect or it is not significant).

The paper is organized as follows. In section \ref{sec:methods} we derive analytical upper bounds using the agnostic or model-free formulation of the learning problem. In connection to the drawbacks pointed in \cite{Friston12} regarding the CV-based inference challenge that ``they are not functions of the complete data'' set, it is worth mentioning that the previous model considers all the available data. Then, the learning algorithm is fitted in the \emph{best} possible way to the empirical data, as shown in section \ref{sec:fitting}, to obtain the \emph{empirical} error. This empirical error represents an upper bound of the \emph{real} (actual) error of the model limited by a deviation quantity that is analytically derived beforehand. Sample size and the empirical settings regarding the complexity of the selected classifiers are key to the proposed neuroimaging methodology as they condition the degrees of freedom and the number of separating functions used to define the aforementioned deviation quantity. In a nutshell, low dimensional scenarios and linear classifiers are required, whenever they easily get a strong link between both errors.  Under these conditions, we can determine which regions across volumes are within these confidence intervals with a probability of $1-\delta$, what corresponds with statistical significance in a group comparison (see section \ref{sec:experiments}). To this purpose, we need to estimate a probability threshold for the obtained accuracy values of each region to reject the null-hypothesis, i.e. $H_0:\pi=0.5$, of the underlying population proportion, thus a regionally specific activation can be stated under the Statistical Agnostic Mapping (SAM) framework.

\section{Methods: bounding the actual error with probability 1-$\delta$}\label{sec:methods}
\subsection{Background on Agnostic learning}
Let assume the agnostic model in the problem of binary pattern classification as proposed in \cite{Haussler92}. Given an independent and identically distributed sample $\mathbf{Z^n}=(\mathbf{Z_1},\ldots,\mathbf{Z_n})$ of $d$-dimensional predictors and classes pairs, $Z_i=(\mathbf{x_i},y_i)$,  where each of them is drawn from the unknown $P\in\mathcal{P}$, the goal is to construct a good approximation to an unknown target function $f^*$, using a class of functions $\mathcal{F}$ : $f:\mathbf{X}\rightarrow U$, and evaluating their goodness by a predefined expected loss: 
\begin{equation}
L_P(f_n) \equiv \mathbb{E}[\ell(y,f_n) |\mathbf{Z^n}] =\int_{\mathbf{X}\times Y} \ell(y,\hat{f_n}) P(d\mathbf{x},dy)
\end{equation}
where the loss function $\ell:Y\times U\rightarrow [0,1]$ and $f_n$ is a random element of the hypothesis space $U$. 

To simplify notation let consider the function composition, i.e. $g=\ell\circ f$, to define the class of functions $\mathcal{G}$ : $g_{\ell}:\mathbf{X}\times Y\rightarrow [0,1]$ with expected loss (probability of error) $P(g_{\ell})$. Thus, the empirical error can be determined by:
\begin{equation}
P_n(g_{\ell})=\frac{1}{n}\sum_{i=1}^ng_{\ell}(\mathbf{Z_i})\leq 1
\end{equation}

A learning algorithm particularly selects $g_n$ given the sample $\mathbf{Z^n}$, i.e. via the empirical risk minimization (ERM) $g_n^*=arg\min\limits_{g\in\mathcal{G}} P_n(g)$ \cite{Vapnik82}, and hopefully provides:
\begin{itemize}
\item a real error $P(g_n)$ (on the ideal infinite population) close to the one obtained on the sample, that is, $P(g_n)\simeq P_n(g_n)$
\item and close to the minimum risk , $L_P^*(\mathcal{G})=\inf\limits_{g\in\mathcal{G}}P(g)=P(g^*)$
\end{itemize}
\subsection{Upper Bound based on concentration inequalities}
Unfortunately, the aforementioned statement of $P(g_n)\simeq P_n(g_n)$
is not generally true. More precisely:
\begin{equation}
P(g_n)>P_n(g_n)+\epsilon>P(g^*)+\epsilon;
\end{equation}
with an arbitrarily $\epsilon>0$. Under the worst case scenario the uniform deviation can be defined as $\Delta_n(Z^n)=\sup\limits_{g\in\mathcal{G}}|P_{n}(g)-P(g)|$, for any $g\in \mathcal{G}$. Using the ERM algorithm we readily get the following concentration inequalities:
\begin{equation}\label{eq:bound1}
\begin{array}{l}
 P(g_n)\leq P(g^*)+2\Delta_n(\mathbf{Z^n})\\
 P(g_n)\leq P_n(g_n)+\Delta_n(\mathbf{Z^n})
\end{array}
\end{equation}
Bounding $\Delta_n(Z^n)$ can be (not readily) achieved by using several theorems and lemmas of the SLT \cite{Vapnik71,Shela72,Sauer72,Massart00,McDiarmid89} to finally get (see Appendix)\footnote{A similar bound can be achieved for the second row in equation \ref{eq:bound1}}:
\begin{equation}\label{eq:bound2}
P(g_n)\leq P(g^*)+8\sqrt{\frac{log(N)}{n}}+\sqrt{\frac{log(1/\delta)}{2n}}
\end{equation}
with probability $1-\delta$, where $N$ is the cardinality of $\mathcal{G}(Z^n)$ or the number of separating functions given the sample realization\footnote{A trivial bound for this quantity can be found: $N\leq 2^n$}. 
           
\section{Fitting the selected function to current data} \label{sec:fitting}
\subsection{Feature extraction and selection}
In order to minimize the left part of equation \ref{eq:bound2} we could minimize one (or both) of the summands on the right. However, they are dependent each other in terms of the classifier complexity \cite{Vapnik82}. One solution could be, as explained in the next section, to prevent the increase of $N\propto\mathcal{O}(n,d)$ given the sample $\mathbf{Z}^n$, by selecting a low classifier order \cite{Cover65}, i.e. a linear decision functions. However, this comes at the cost of a maybe non-negligible empirical error.

As an attempt to reduce the ratio $d/n$ (\emph{curse of dimensionality}), the machine learning community usually tends to employ feature extraction and selection (FES) methods to enhance the classification performance while preserving the system complexity. This can be achieved by removing irrelevant features from the sample, which can also facilitate interpretation (FS), and by identifying multivariate sets of meaningful features (FE) that best discriminate the classes \cite{Rondina15}. The final aim is to provide an almost linearly separable classification problem in the \emph{feature space}. 

Several methods have been employed in neuroimaging aiming at reducing the dimensionality ($d$) of the problem (in relation to $N$) based on statistical tests for FS \cite{DeMartimo08}, matrix decompositions \cite{Mcintosh96} or even deep learning architectures for FES \cite{Martinez19}. To validate the methodology proposed in this paper, we perform FE using a popular method in neuroscience, such as the Partial Least Squares (PLS) algorithm \cite{Mcintosh96}. PLS methods have been demonstrated its utility in describing the relation between brain activity and experimental design or behaviour measures within a multivariate framework (see \cite{Mcintosh96,Rosipal06,Gorriz18} and the appendix for mathematical details and the interpretation of the PLS-maps as a classical t-test).

\subsection{Linear Decision functions: a small upper bound}

Regularized linear decision functions have been recently applied to neuroimaging for detecting activation patterns, and compared to parametric hypothesis testing, such as univariate t-tests \cite{Mouro-Miranda05,Gomez19,Khundrakpam15}. In general, they have limited their analyses to provide in-sample estimates based on resampling, failing to demonstrate their out-of-sample performance in terms of confidence intervals. 

As stated before, the minimization of the left part of inequality \ref{eq:bound2} can be achieved by decreasing the number of separating functions $N$ given the sample ($\mathbf{Z^n}$). This quantity is indeed decreased by selecting a linear decision function-based classifier in a binary classification problem, following the results in the extant literature \cite{Cover65,Vapnik82}, etc. After transforming and selecting the feature set by FES methods, the concentration inequalities \ref{eq:bound2} obtained with linear classifiers result in a strong association with a given confidence level whenever the extracted features are significant across regions of interest (ROIs) and group comparisons.

\begin{figure*}
\centering
\includegraphics[width=\textwidth]{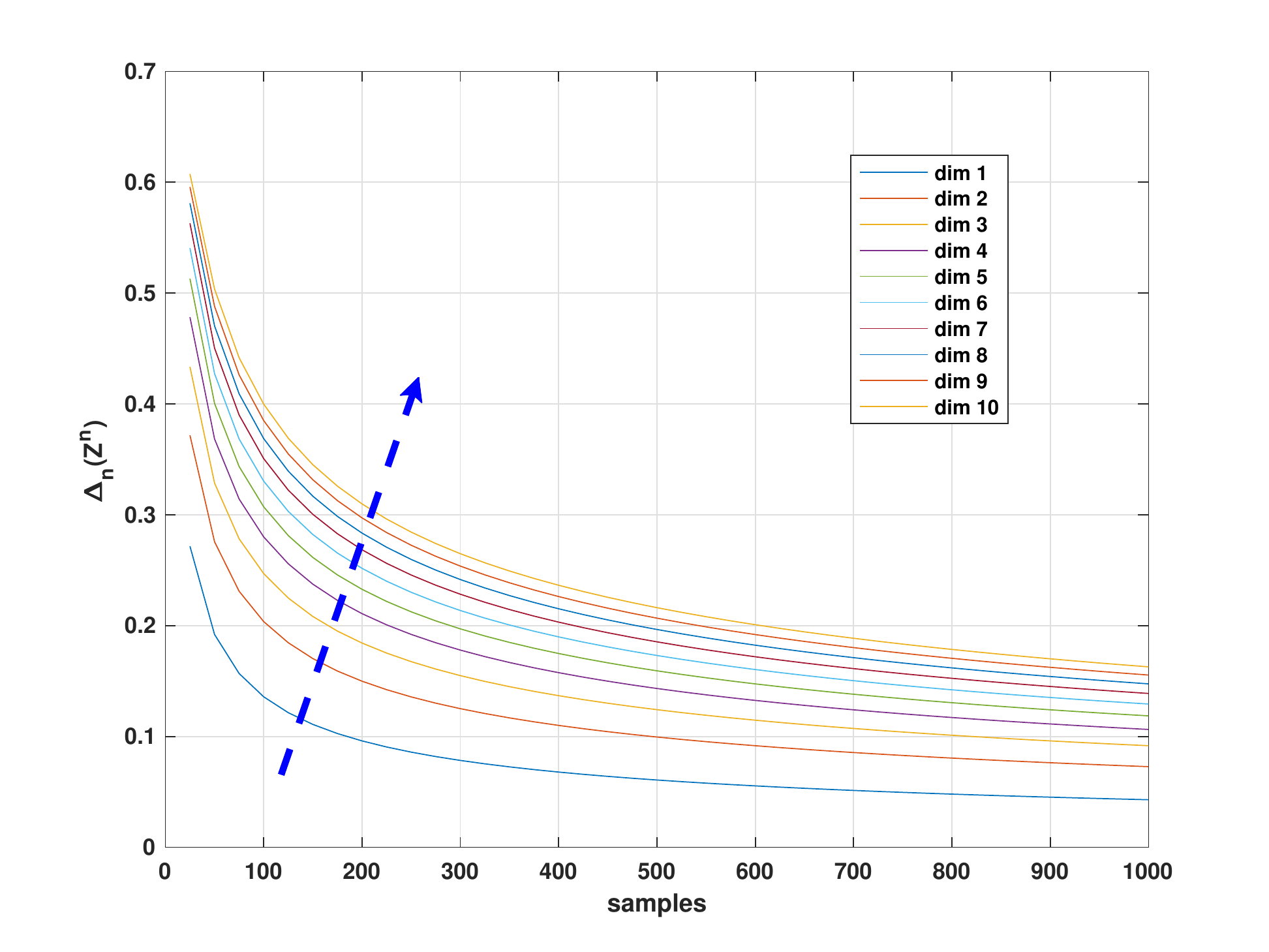}
\caption{The upper bound of inequality \ref{eq:bound2_exp} connecting actual and empirical errors with $95\%$ level of confidence. Note how increasing the feature dimension results in a larger bound (blue solid line). However, working in low dimensional scenarios, i.e. $d=1,2$, and using medium-sized datasets ($n=500$), the confidence interval is less than $10\%$.}
\label{fig:bound}
\end{figure*}

Beyond the existing caveats and solutions when using regularization methods in neuroimaging for FS, we adopt the linear support vector machine (SVM) classification algorithm which allows us to tentatively evaluate the worst case of $N$, that is, $\mathbb{S}_n(\mathcal{G}) \equiv \sup\limits_{\mathbf{x^n}\in\mathbf{Z^n}}(N)$ and to set the following upper bound \cite{Vapnik82}:
\begin{equation}\label{eq:bound1_exp}
\Delta_n(\mathbf{Z^n})\leq \sqrt{\frac{h((\log(2n/h)+1)-log(\delta/4)}{n}}
\end{equation}
with probability $1-\delta$ and $h$ is the VC dimension, e.g $h=d+1 $ for linear classifiers. In the same manner, several upper bounds could be tested based on several innovative concepts and paradigms, such as the ones based on data distributions, set's shape, Rademacher averages, pseudo-dimension, fat-shattered dimension, etc. \cite{Antos02,Vidyasagar03}. We preferred to use, due to its simplicity, the upper bound recently proposed in \cite{Gorriz19}. The latter is strongly grounded on the geometrical assumption of \emph{in general position} distributed samples and the function-counting theorem of homogeneously linearly separable dichotomies \cite{Cover65}:
\begin{equation}\label{eq:bound2_exp}
\Delta_n(\mathbf{Z^n}) \leq 
\sqrt{ \frac{ (d-1)\log(n+1)+(2+\log(1/\delta)) }{2n}}
\end{equation}
With the help of expressions such as inequalities \ref{eq:bound1_exp} and \ref{eq:bound2_exp}, we can even evaluate the deviation of the empirical error from the actual error at voxel level, although it is preferable, for the aforementioned reasons, to do it region-wise using a fitted linear SVM classifier in the multivariate feature space (see figure \ref{fig:bound}). In this sense, the motivation for a multivariate framework in assessing the areas of relevance is analogous to other proposed techniques for addressing the multiple comparison problem in functional imaging, e.g. Random Field Theory for neuroimaging analysis \cite{Frackowiak04} or the classical p-value corrections for multiple comparison after null-hypothesis testing. In general, only those voxels (or ROIs) showing a tight association, i.e. high performance in terms of accuracy, should be considered as relevant maps or patterns in that particular condition with probability $1-\delta$.

\section{Statistical Agnostic Mapping}
The significant areas derived from SAM correspond by construction with those regions having an empirical error $P_n(g_n)$ that, under the worst case scenario, has associated an actual error $P(g_n)$ greater than the random guess accuracy $\pi=0.5$. Confidence intervals derived from the concentration inequalities allow us to bound the worst case at the ``upper'' border of the confidence interval, providing a protective inference. Thus, within this confidence interval, a significance test can be used to make an inference about whether the accuracy value for a specific region differs from the null-hypothesis of the random proportion $\pi=0.5$ (see Appendix). Therefore, the statistical significance of any region is assessed, in combination with confidence intervals, by evaluating the p-value of any ROI at a given significance level, i.e. $\alpha=0.05$. A total of $l=116$ standardized regions \cite{Tzourio02} were analysed within a protective interval, avoiding the limitations of significant tests to distinguish statistical from practical importances (see Appendix).

\begin{figure*}
\centering
\includegraphics[width=\textwidth]{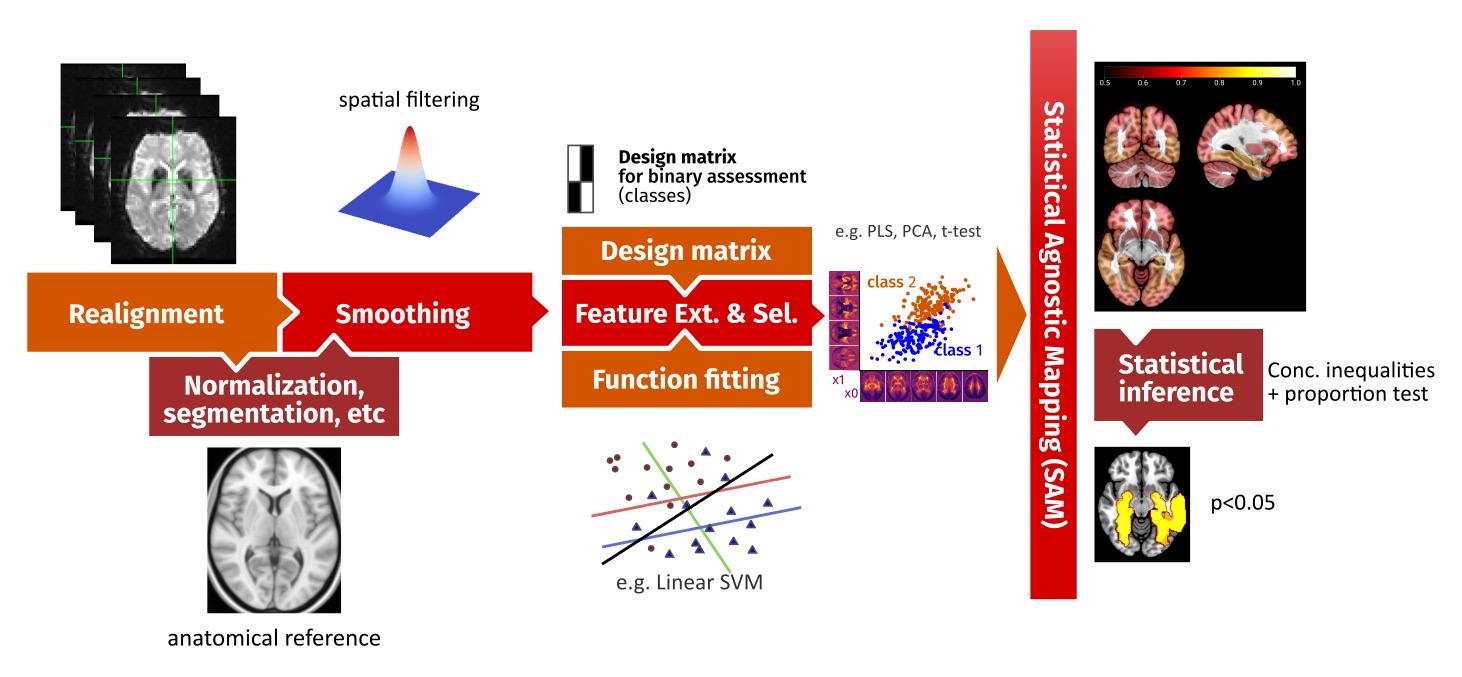}
\caption{Complete diagram of the proposed methodology including typical preprocessing steps in SPM for different modalities (left column of blocks), classification fitting and FESfor actual risk estimation  (middle column)  and inference to derive the SAM (right column). }
\label{fig:diagram}
\end{figure*}

In the following sections we will show how the combination of the aforementioned protective intervals and significance tests may be used to derive a SAM in different group comparisons, such as Alzheimer's disease (AD) vs healthy controls (HC), Parkinson's disease (PD) vs HC and on a well-known example of single-subject activation map in fMRI, and how they relate with the classical approach based on null-hypothesis testing, i.e. two sample t-test with corrected-p value. Unlike, previous approaches, the proposed model-free method is less specific but more robust against sample size, artifacts and nuisance effects. See the complete diagram of the poposed method in figure \ref{fig:diagram}.

\section{Experiments}\label{sec:experiments}

The aim of this section is to present a novel methodology in neuroimaging based on analytical concentration inequalities, and to experimentally compare them to the accepted framework used by the neuroscience community based on the SPM analysis \cite{Friston95}. Thus, we will assess several experiments collected from well-known databases that include imaging data from patients with a variety of conditions/pathologies.  Nevertheless, we will avoid somewhat related theoretical discussions about the comparison of both branches of statistics, referring the readers to the introduction section in this paper and the vast extant literature addressing these issues \cite{Friston2013,Friston12,Lindquist13, Reiss15}.

All the datasets were preprocessed using standardised neuroimaging methods and protocols implemented by the SPM software (registration in MNI space, spatial normalization and segmented to differentiate brain tissues, e.g. Grey matter (GM)) \cite{Friston95}. For further comparison with the SAM proposed in this paper, significance maps were obtained with SPM using a standard two-sample t-test with FWE p-value$=0.05$ (and null extent threshold -voxels-). We first conducted a first-level analysis to derive the GLM for the dataset under assessment (a design matrix for group comparisons) and then, in the 2nd-level analysis, the contrast images were fed into a GLM for implementing the statistical test.

\begin{table}[htbp]
   \centering
   \caption{Demographics details of the datasets (ADNI, PPMI, VV and fMRI), with group means with their standard deviation}
   \label{tab:demog}
   \begin{tabular}{lccccc}
     \hline
     & Status & Number &	Age	& Gender (M/F) &	MMSE\\ \midrule
MRI  \\
ADNI& NC	        &   229    &	  75.97$\pm$5.0	    &   119/110	  & 29.00$\pm$1.0 \\
&MCI          &	  401   &	  74.85$\pm$7.4	    &   258/143	  & 27.01$\pm$1.8  \\
& AD	        &  188	   &     75.36$\pm$7.5 	&     99/89	  & 23.28$\pm$2.0  \\\midrule
SPECT 	\\
PPMI& NC	        &   194    &	  53.02$\pm$2.27	    &   129/65	  & --     \\
& PD	        &   168    &	  53.14$\pm$2.37	    &   103/65	  & --     \\\midrule
SPECT \\
VV & NC	        &   108    &	  69.05$\pm$14.53	    &   54/54	  & --     \\
& PS	        &   100    &	  68.62$\pm$13.41	    &   53/47	  & --    \\
\midrule
fMRI \\
Auditory & Res	        &   41    &	 -    &  1  & --     \\
& List        &   43    &	  -	    &   1	  & --    \\ \bottomrule
\end{tabular}
\end{table}

\subsection{A structural MRI (sMRI) study: the ADNI database}\label{sec:exp1}

Data used in preparation of this paper were obtained from the Alzheimer's Disease Neuroimaging Initiative (ADNI) database (adni.loni.usc.edu). The ADNI was launched in 2003 by the National Institute on Aging (NIA), the National Institute of Biomedical Imaging and Bioengineering (NIBIB), the Food and Drug Administration (FDA), private pharmaceutical companies and non-profit organizations, as a 60 million dollars, 5-year public-private partnership. The primary goal of ADNI has been to test whether serial magnetic resonance imaging (MRI), positron emission tomography (PET), other biological markers, and clinical and neuropsychological assessment can be combined to characterise the progression of mild cognitive impairment (MCI) and early AD. The construction of sensitive and specific biomarkers of very early AD progression is intended to aid researchers and clinicians to develop new treatments and monitor their effectiveness, as well as lessen the time and cost of clinical trials.

The Principal Investigator of this initiative is Michael W. Weiner, MD, VA Medical Center and University of California San Francisco. ADNI is the result of efforts of many co-investigators from a broad range of academic institutions and private corporations, and subjects have been recruited from over 50 sites across the U.S. and Canada. The initial goal of ADNI was to recruit 800 subjects but ADNI has been followed by ADNI-GO and ADNI-2. To date these three protocols have recruited over 1500 adults, ages 55–90, including cognitively normal older individuals, people with early or late MCI, and people with early AD. The follow up duration of each group is specified in the protocols for ADNI-1, ADNI-2 and ADNI-GO. Subjects originally recruited for ADNI-1 and ADNI-GO had the option to be followed in ADNI-2. For up-to-date information, see www.adni-info.org.

The ADNI database contains 1.5 T and 3.0 T t1w MRI scans for AD, MCI, and cognitively normal controls (NC) which are acquired at multiple time points. Here we only included 1.5T sMRI corresponding to the three different groups of subjects: NC, AD and MCI. The original database contained more than 1000 T1-weighted MRI images, comprising 229 NC, 401 MCI (312 stable MCI and 86 progressive MCI) and 188 AD. Although for the proposed study, only the first medical examination of each subject is considered, resulting in 818 GM images. Following the recommendation of the National Institute on Aging and the Alzheimer's Association (NIA-AA) for the use of imaging biomarkers \cite{NIA18}, we considered the group comparison $NC vs AD$ for establishing a clear framework for comparing statistical paradigms (SPM and SAM). Thus, the MCI class is strictly based on clinical criteria, without including any other biomarker \cite{McKhann11}. Demographic data of subjects in the database is summarized in Table \ref{tab:demog}.

\subsubsection{Classification Results}

\begin{figure*}
\centering
\includegraphics[width=\textwidth]{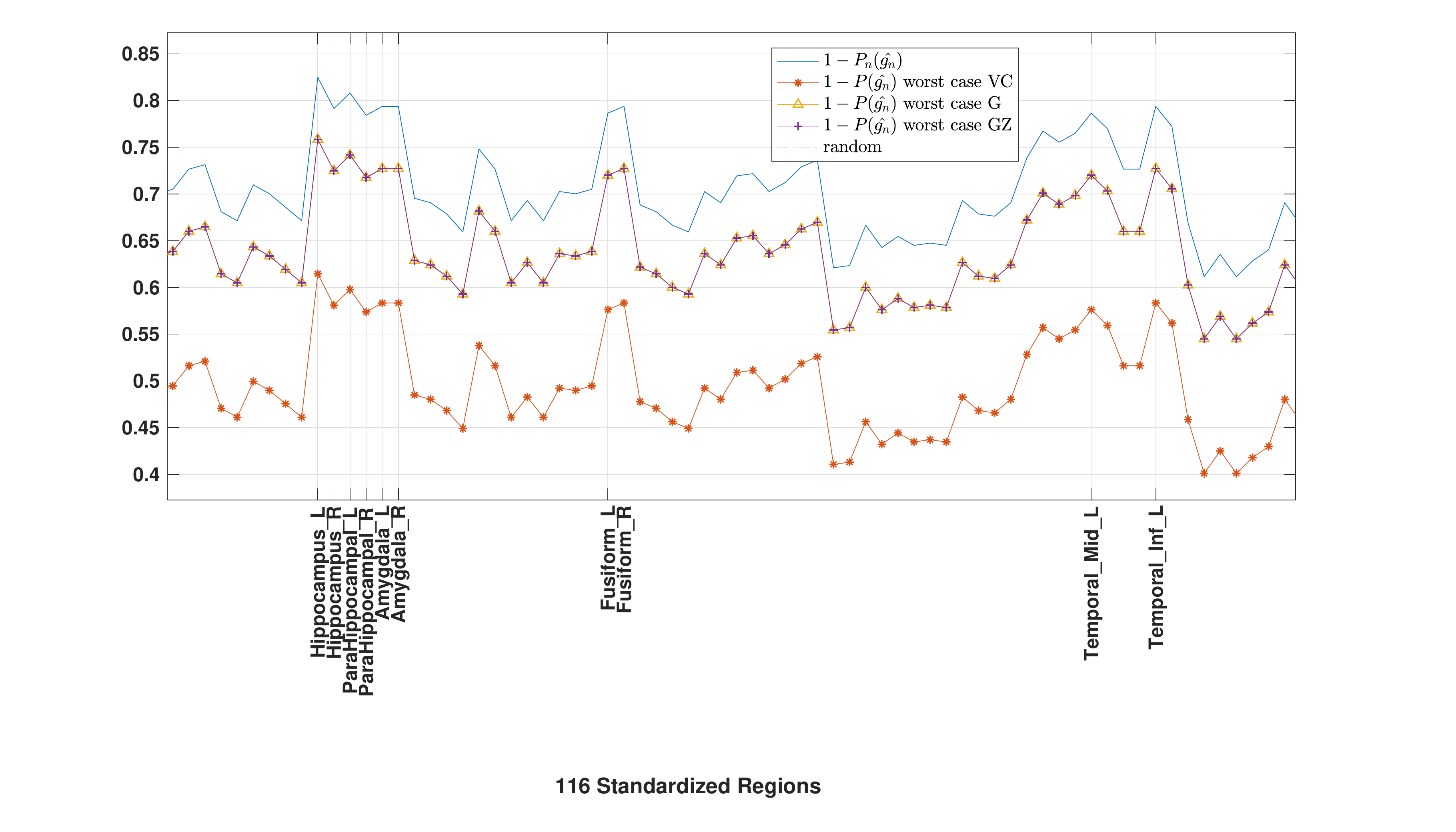}
\caption{Accuracy values and upper bounds in standardized ROIs (only significant regions from $\#30$ to $\#90$ are shown) for three methods based on concentration inequalities in \ref{eq:bound1} and \ref{eq:bound2}. We highlight several regions, relevant in the biological definition of AD, i.e. Hippocampus, Temporal, Amygdala and Parahippocampal regions, corresponding to peaks of these curves. Moreover, observe how the VC approach is more pessimistic than the one based on \cite{Gorriz19}. The confidence interval is drawn in the space between the solid blue line and the colored lines.}
\label{fig:class1}
\end{figure*}

First, the proposed methodology try to fit in an optimal way a linear SVM classifier in the feature space obtained after a FES approach (PLS).  With the aim of applying a regression-type analysis to the dataset, we parcellated the brain volume into 116 standardized regions \cite{Tzourio02} and then, obtained an optimistic estimation of the actual error $P(g_n)$ as shown in solid blue line in figure \ref{fig:class1}. This estimation is corrected by the use of upper bounds drawing a novel set of accuracy values (proportions) and a confidence interval, depending on the selected theoretical method, i.e. Vapnik's bound.  The lower accuracies in this plot corresponds to the worst cases as considered by the selected concentration inequalities.

It is worth mentioning that the results, shown in figure \ref{fig:class1}, are obtained with the first PLS component extracted by this regression analysis ($d=1$). This PLS score for each subject can be conceptualised as the representation of the subject into a multi-dimensional reference system as described in \cite{Gorriz18} (see supplementary material). 

\subsubsection{Statistical Agnostic Maps}

\begin{figure*}
\centering
\includegraphics[width=\textwidth]{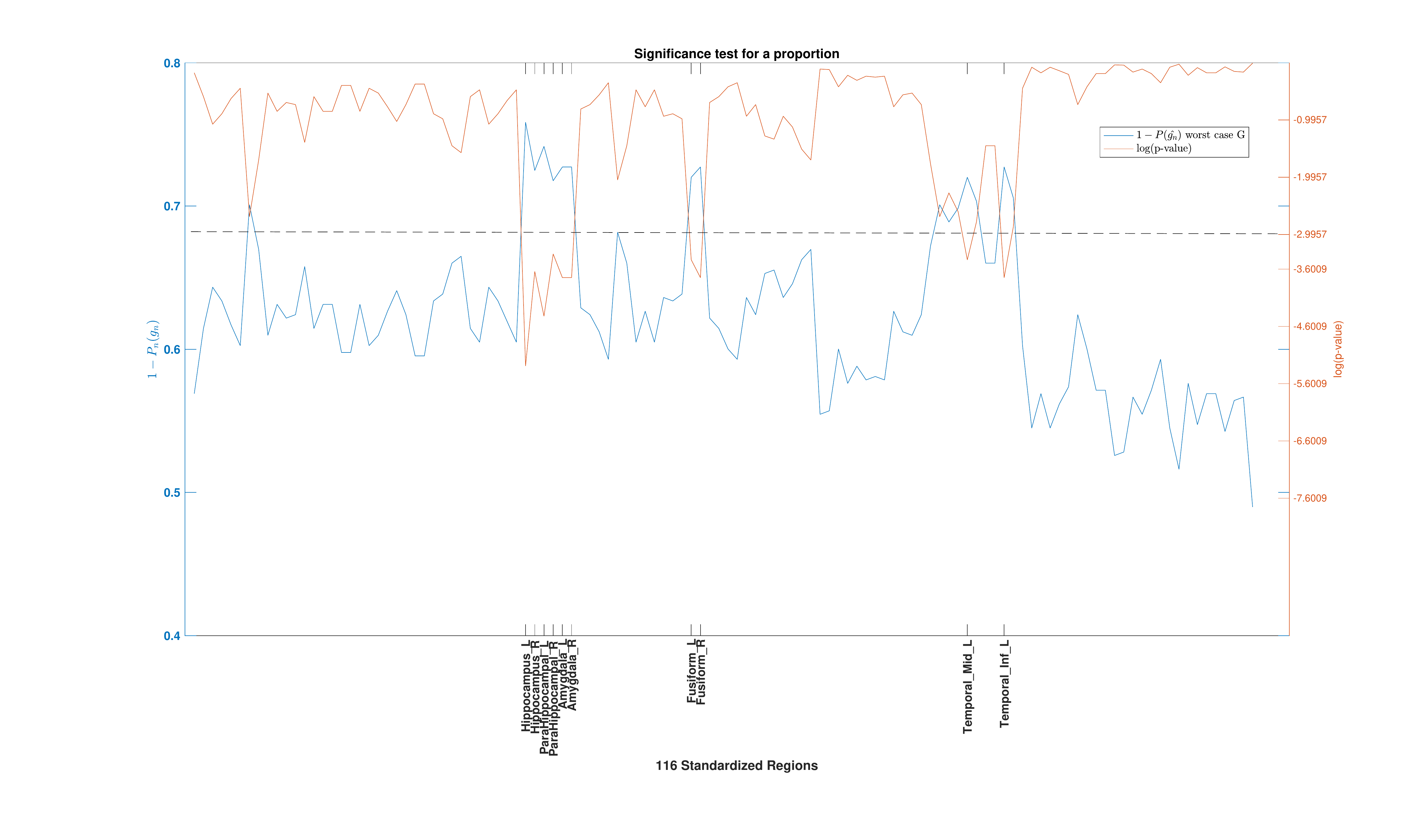}
\caption{Accuracy values in the worst case using the method in \cite{Gorriz19} and the set of probabilities (log(p-values)) within the confidence interval. The ROIs (p$<0.05$) are detected out of $116$ standardized regions using a significance test for a proportion $\pi$ (see appendix). Note that we show the probability of observation (in the right ``y axis'') of the set of accuracy values under $H_0$, i.e. random distribution.}
\label{fig:SAM1}
\end{figure*}

In the aforementioned figures we heuristically identified all those relevant regions for the characterisation of AD based on absolute values. Therefore, a definition of relevancy in terms of hypothesis testing within confidence intervals is required. Following the method presented in the appendix, we provide an automatic (and statistically elegant) method for selecting ROIs in which a regionally specific activation is identified. As depicted in figure \ref{fig:SAM1}, the main result is the SAM obtained with the same p-value as the one of confidence intervals using concentration inequalities. 

\begin{figure*}
\centering
\includegraphics[width=0.49\textwidth]{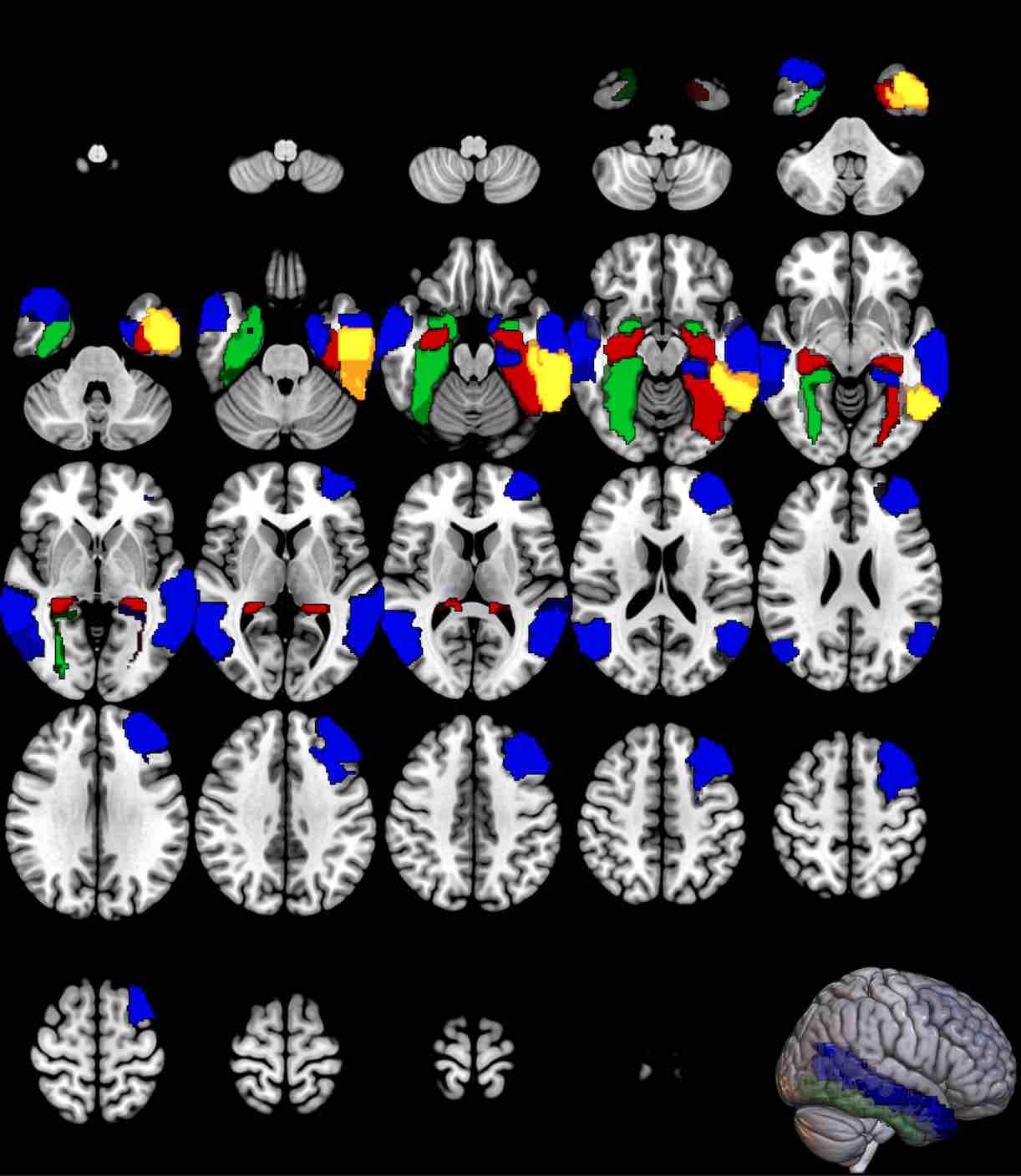}
\includegraphics[width=0.49\textwidth]{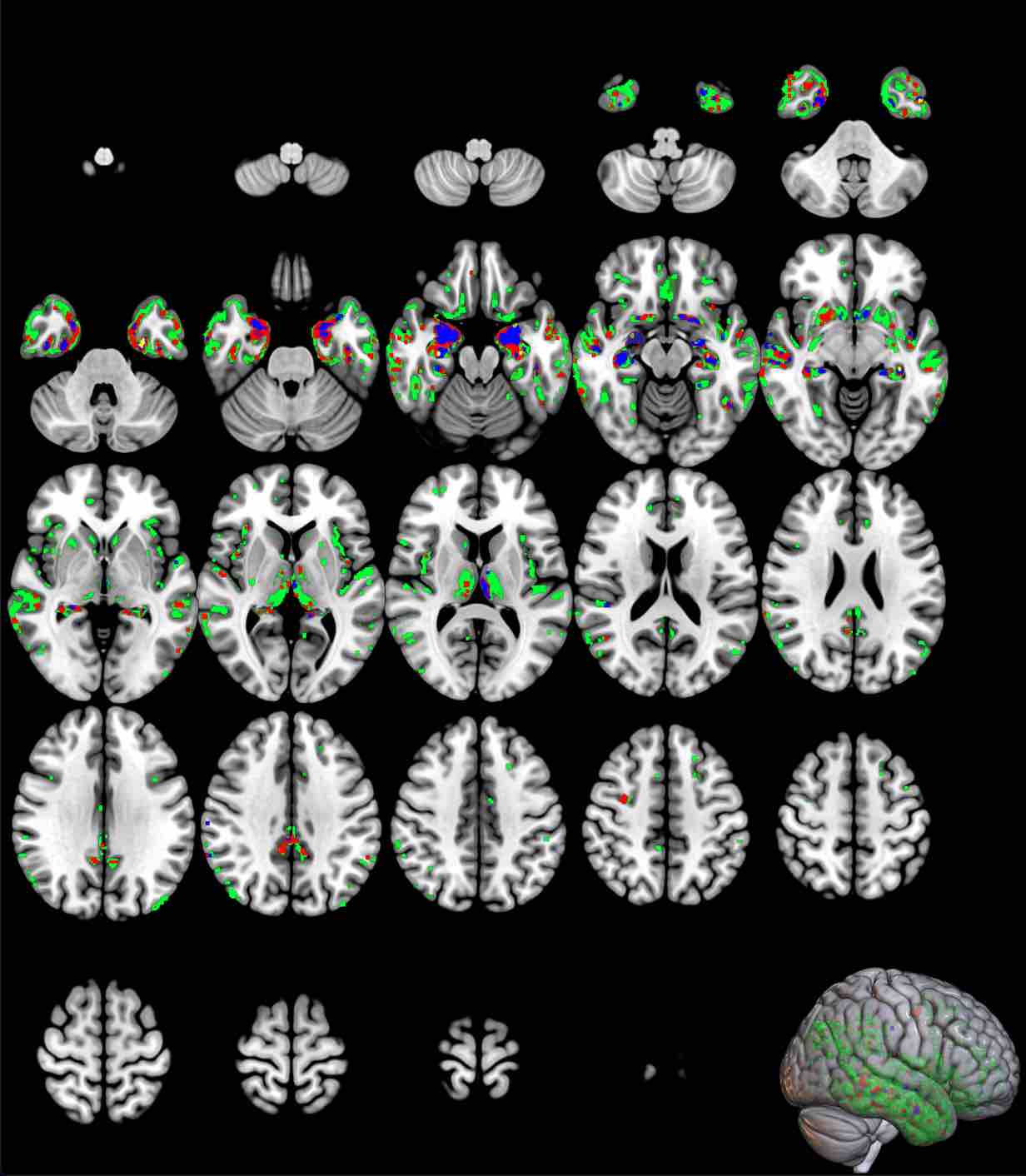}
\caption{Statistical comparison of brain volumes using SAM (left) and SPM (right) in the ADNI database. Green area corresponds to the whole dataset while the rest of colors (red, blue, yellow) are linked to data subsets, which are plotted in increasing $n$ (opacity of representations is preserved for clarity reasons). The ROIs selected for increase $n=50, 100, 200, 417$, satisfy $S_{j} \subset S_{j+1}$ except for $n=50$ where an additional region ``Frontal Mid L'' is selected. It is worth mentioning that all the ROIs extracted in different sample-size configurations were included in the confidence interval and with probability close to the significance level ($\alpha=0.05$).}
\label{fig:SAM2}
\end{figure*}

\begin{figure*}
\centering
\begin{subfigure}{0.49\textwidth}
\includegraphics[width=\textwidth]{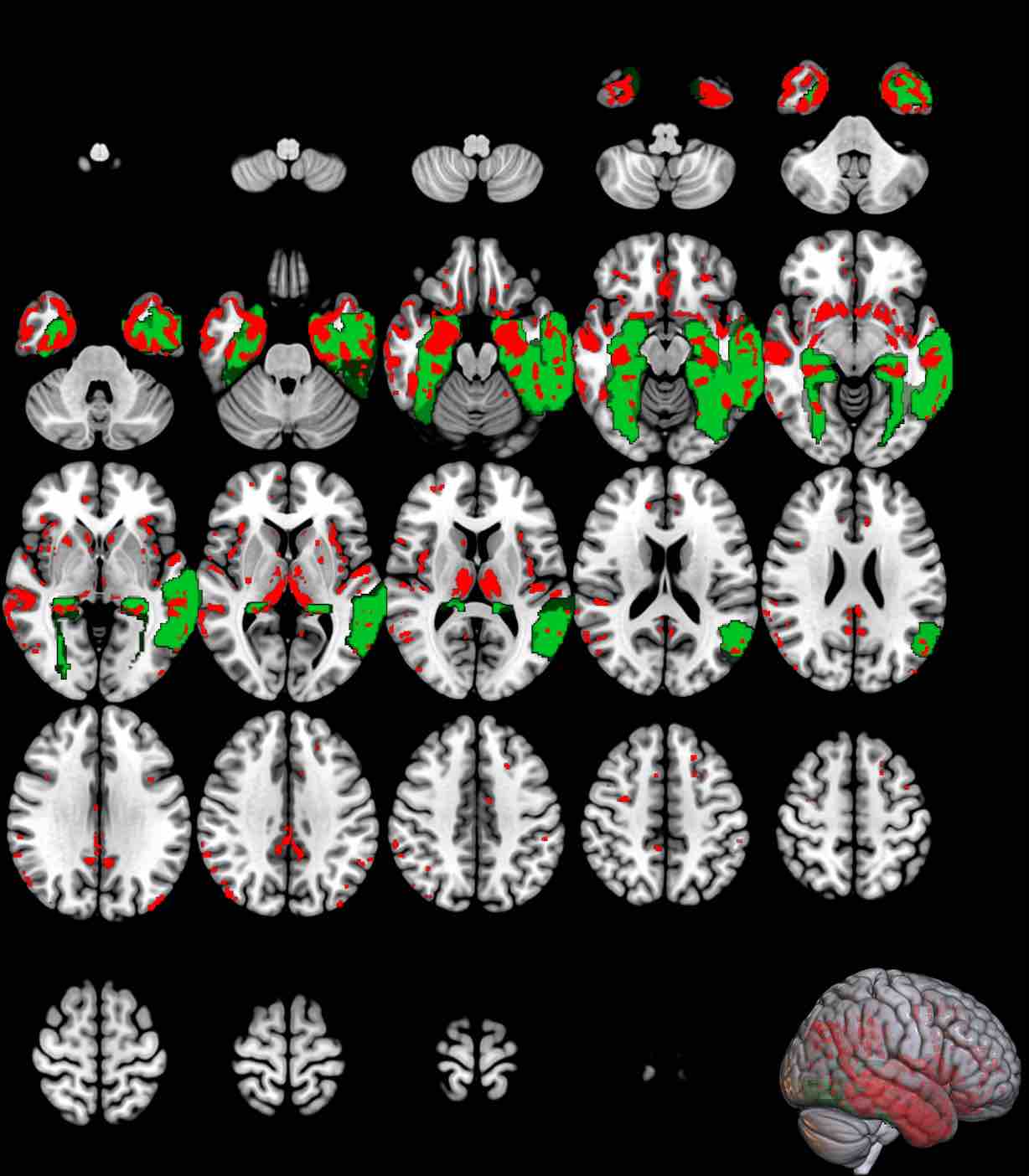}
\caption{}
\end{subfigure}
\begin{subfigure}{0.49\textwidth}
\centering
\includegraphics[width=\textwidth]{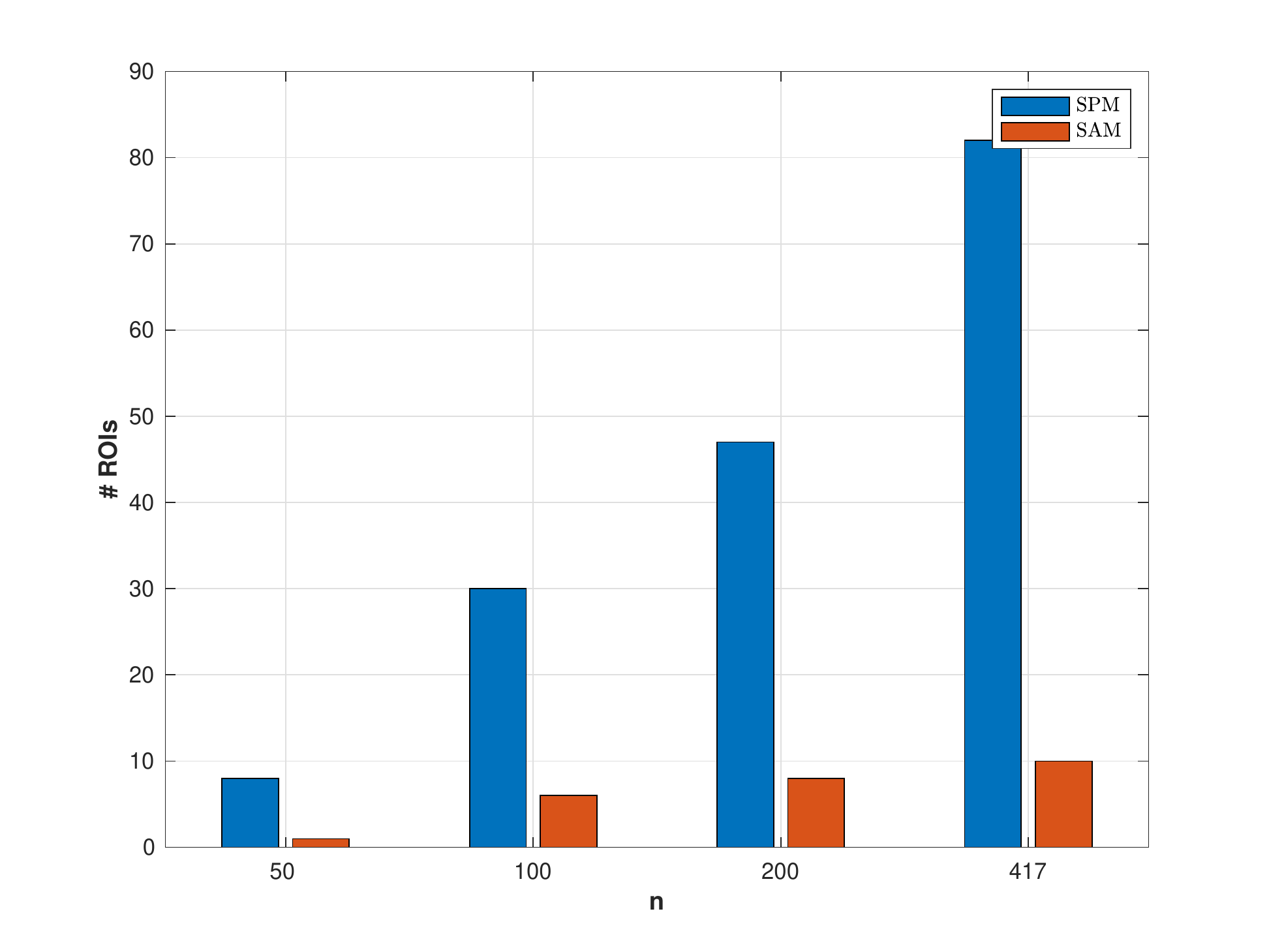}
\includegraphics[width=\textwidth]{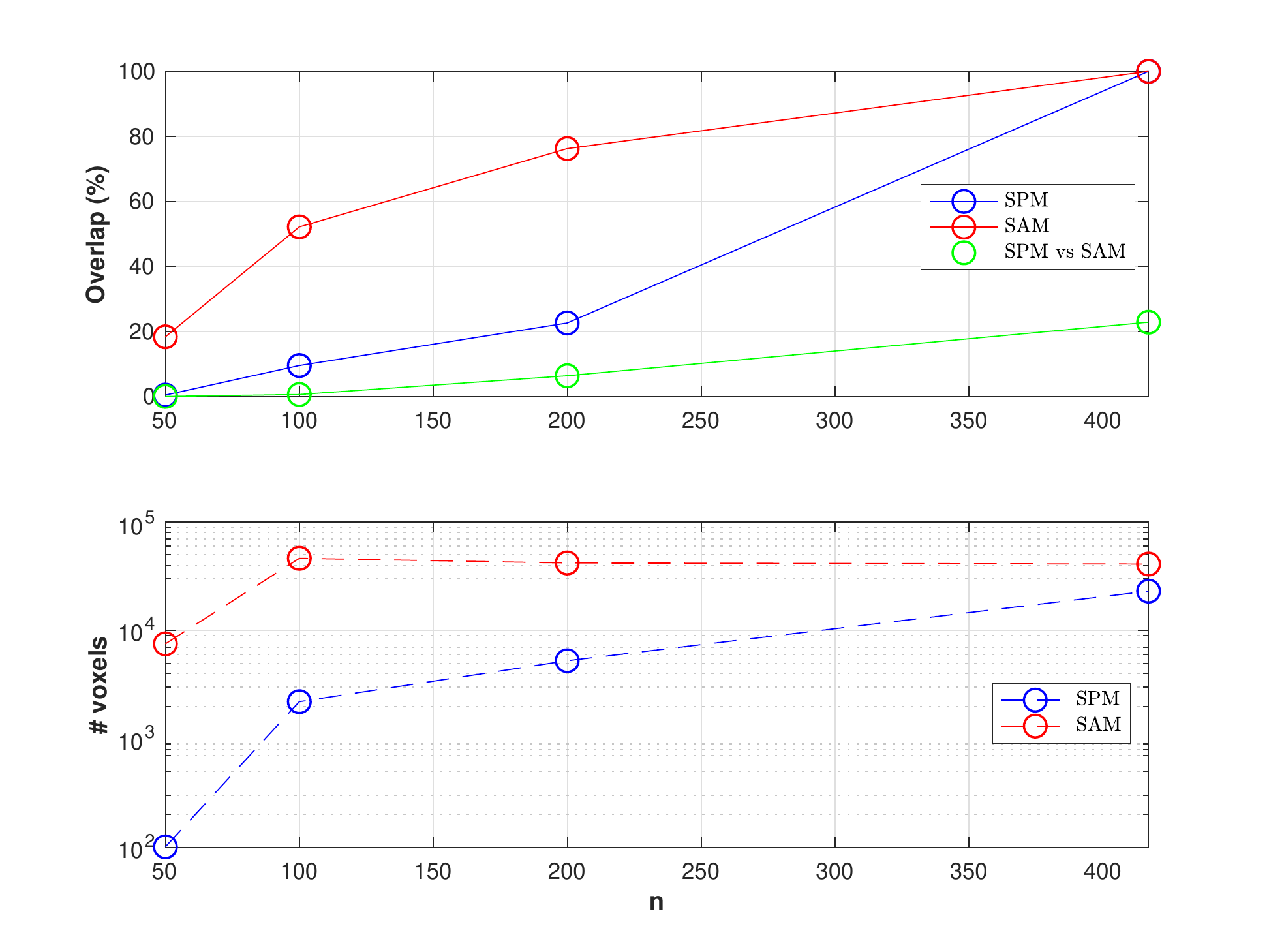}
\caption{}
\end{subfigure}
\caption{(a) SPM (red) over SAM (green) using the complete ADNI dataset ($n=417$). (b) overlap analysis vs sample size. Observe how the SPM activation map linearly increases with $n$ and is located on more than $80$ standardized regions with the whole dataset (although part of these isolated activation voxels could be removed from the map using the extent threshold.)}
\label{fig:SAM3}
\end{figure*}

Finally, a direct comparison with the SPM approach is shown in figures \ref{fig:SAM2} and \ref{fig:SAM3}, in terms of the sample-size analysis and the relevant regions determined by both methods. Key to this comparison is the different working operations, i.e. SAM includes the spatial structure of data at the first FES stage, whilst SPM do it at the final stage, by means of RFT. For this reason, SPM is more specific (voxel-wise) but widespread comparing to SAM. The number of identified ROIs conforming the SPM increases as the number of sample increases, unlike the proposed approach, which provides the same volumetric differences for $n=200, 417$. It is worth mentioning that from the perspective of SLT, due the small ratio $n/d$ in all these experiments proposed in this paper (and in the extant literature), we are dealing with the ``small sample size problem''. In terms of classical statistics (SPM) this derives in a challenging scenario that constrains the generalisation of the results from small datasets to new unseen samples.

Figure \ref{fig:SAM3} shows that main regions identified by SPM are included in the ROIs deployed by SAM-based approach. In addition, the number of ``activated'' voxels in SPM is associated with sample size and these voxels are widespread across several anatomical regions.  The number of voxels in ROIs obtained by SAM is almost independent on the sample size, except for the extreme case $n=50$, an given the magnitude of the effect being sought in the HCvsAD comparison.

\subsection{A SPECT study: the PPMI database}\label{sec:exp2}

Data used in the preparation of this article were obtained from the Parkinson's Progression Markers Initiative (PPMI) database (www.ppmi-info.org/data). For up-to-date information on the study, visit www.ppmi-info.org. PPMI is a public-private partnership funded by the Michael J. Fox Foundation for Parkinson's Research and funding partners, including all partners listed on www.ppmi-info.org/fundingpartners.

Informed consents to clinical testing and neuroimaging prior to participation of the PPMI cohort were obtained, approved by the institutional review boards (IRB) of all participating institutions. The PPMI obtained written informed consent from all study participants before enrolled in the Initiative. None of the participants were taking any PD medication when they enrolled in the PPMI.

The inclusion criteria adopted in the PPMI cohort study are available in http://www.ppmi-info.org/wp-content/uploads/2014/06/PPMI-Amendment-8-Protocol.pdf. This diagnostic procedure also includes a confirmation step based on imaging biomarkers. A selection of $N = 269$ DaTSCAN images from this database were used in the preparation of the article. Specifically, the baseline acquisition from $158$ subjects suffering from PD and $111$ NC was used. In addition, a similar SPECT image database from a the ``Virgen de la Victoria'' (VV) Hospital (Malaga, Spain) was used to validate and generalise our findings to a dataset that contains a more complex pattern in the Parkinsonian Syndome (PS) class derived from a clinical diagnosis criteria \cite{Illan12}  (see table \ref{tab:demog}). 

\subsubsection{Effect size in classification}

Following the methodology presented in the latter section we will show i) the robustness of the proposed methodology in limited sample sizes regarding effect size and ii) a quantitative interpretation of effect size appealing to image classification in diagnostics. As already commented in \cite{Button13}, studies with low statistical power require large effects to be observed by hypothesis testing with a pre-specified p‑value threshold (typically 0.05). In DatSCAN imaging of PD the true effect size is known to be considerably large on specific regions, e.g. striatum.  On the contrary, large effects observed in studies with reduced sizes do not assure that the true effect is large, or even that it exist at all. These studies are usually related to poorly mechanistically grounded hypothesis \cite{Friston12} or a bad specification of clinical analysis plans to conform the set of observations, i.e. dataset \cite{Button13}.

\begin{figure*}
\centering
\includegraphics[width=1\textwidth]{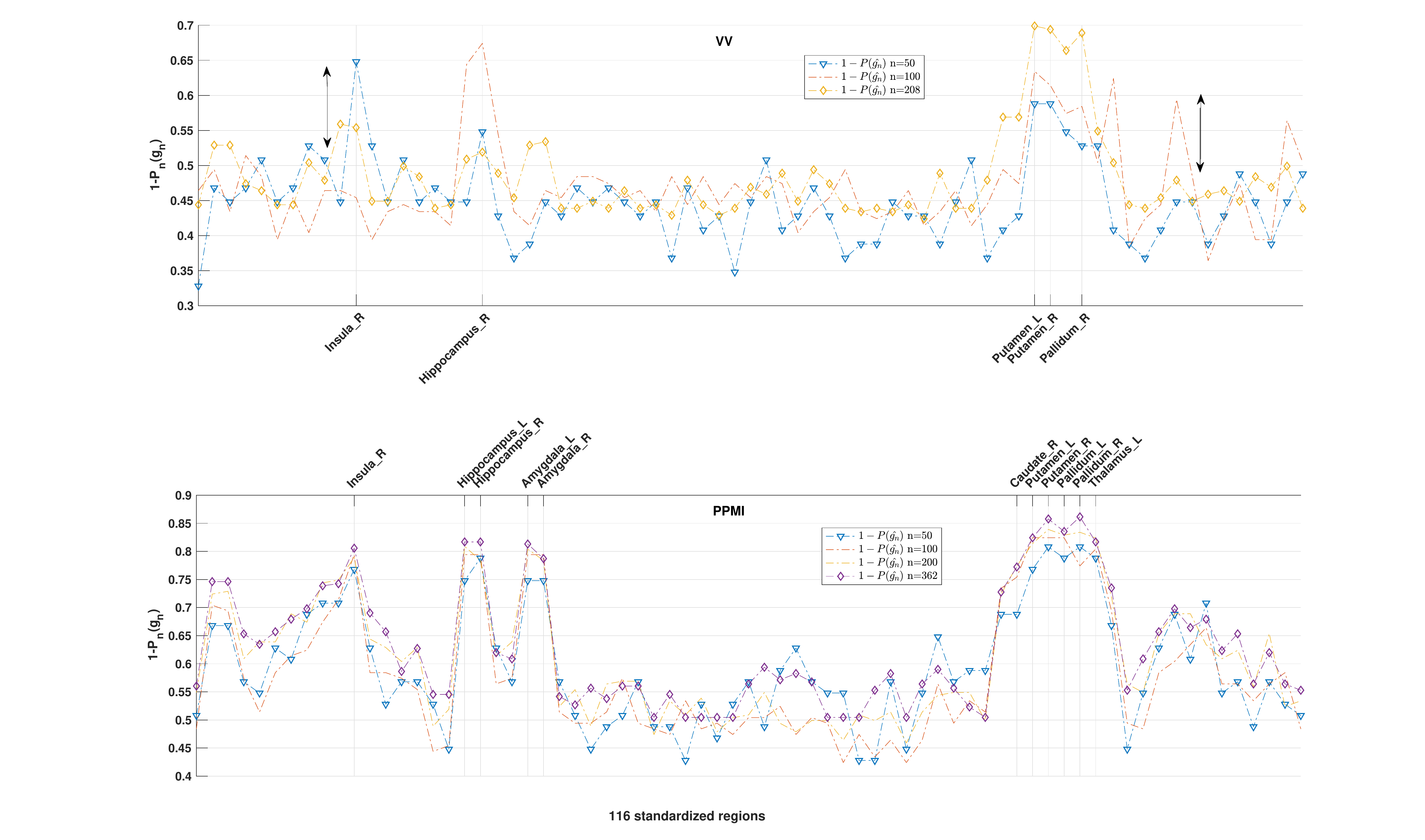}
\caption{Effect sizes and significant ROI selection using the proposed methodology. Observe in top figure (VV) the black arrows highlighting the observed trivial effects, outside the specific region, in studies with low statistical power, i.e. $n=50$. In addition, we also remark the reduction of ``true effect'' in the top figure compared to the same effect depicted in the bottom figure (PPMI), due to the presence of a more complex PD-plus pattern in the VV dataset.}
\label{fig:class3}
\end{figure*}

These issues are can be observed in figure \ref{fig:class3}, where accuracy values are shown for increasing $n=50,100,\ldots$. Effect sizes are large when they can discriminate between subjects that do and do not show an effect \cite{Friston12}. Large (but trivial under our hypothesis PDvsHC) effects observed for $n=50,100$ samples reduce as the sample size increases in the VV dataset, unlike in the PPMI dataset. In the latter dataset, the proposed methodology provides almost the same accuracy values, which are, in general, shifted up w.r.t the former database, for a wide range of samples sizes of randomly selected subjects. Anyway, our method reports effect sizes (in terms of accuracy values) and confidence intervals alongside exact p values, thus improving the strength of inference.  

\subsubsection{Statistical Agnostic Maps}

\begin{figure*}
\centering
\begin{subfigure}{0.49\textwidth}
\centering
\includegraphics[width=\textwidth]{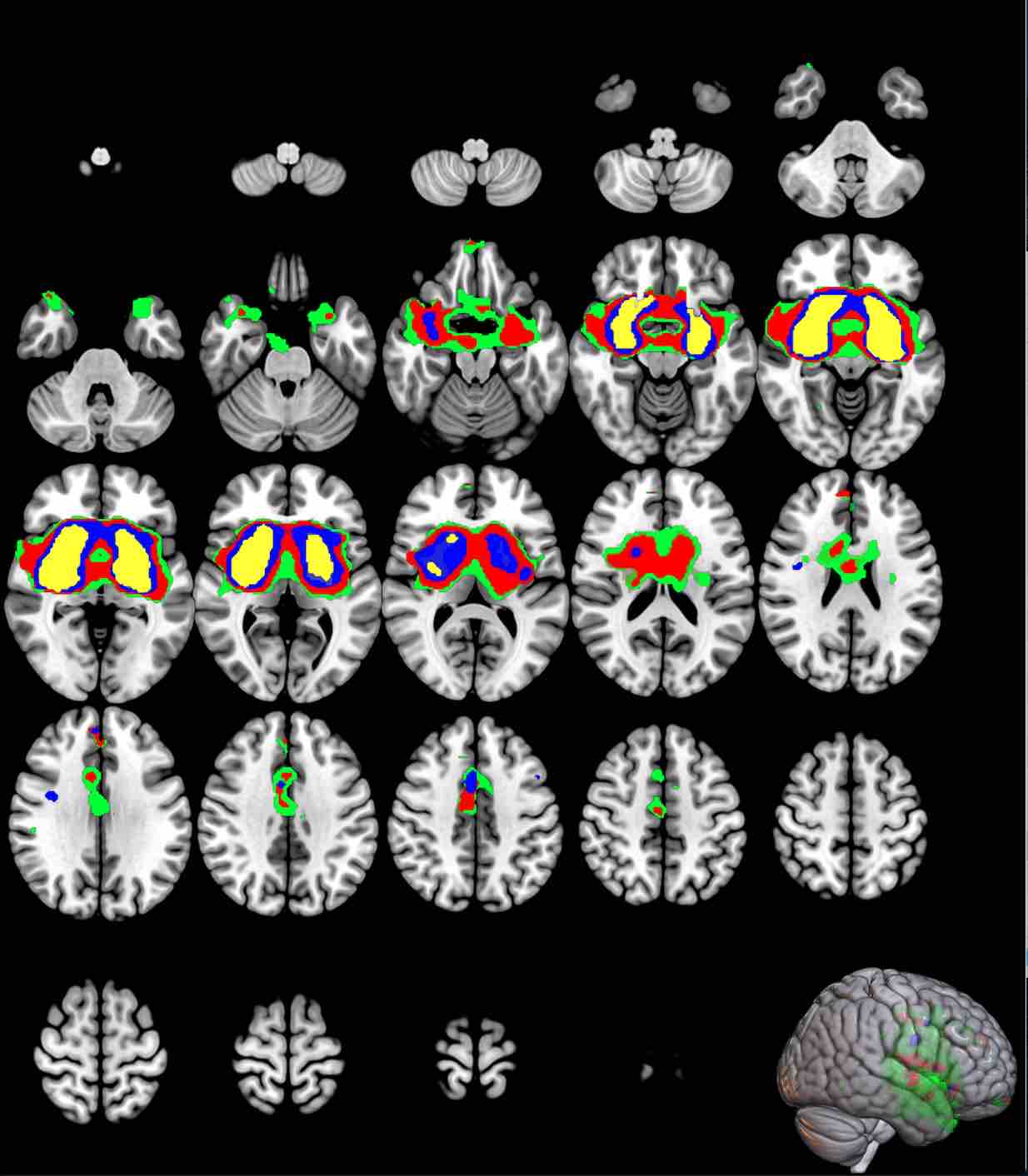}
\includegraphics[width=\textwidth]{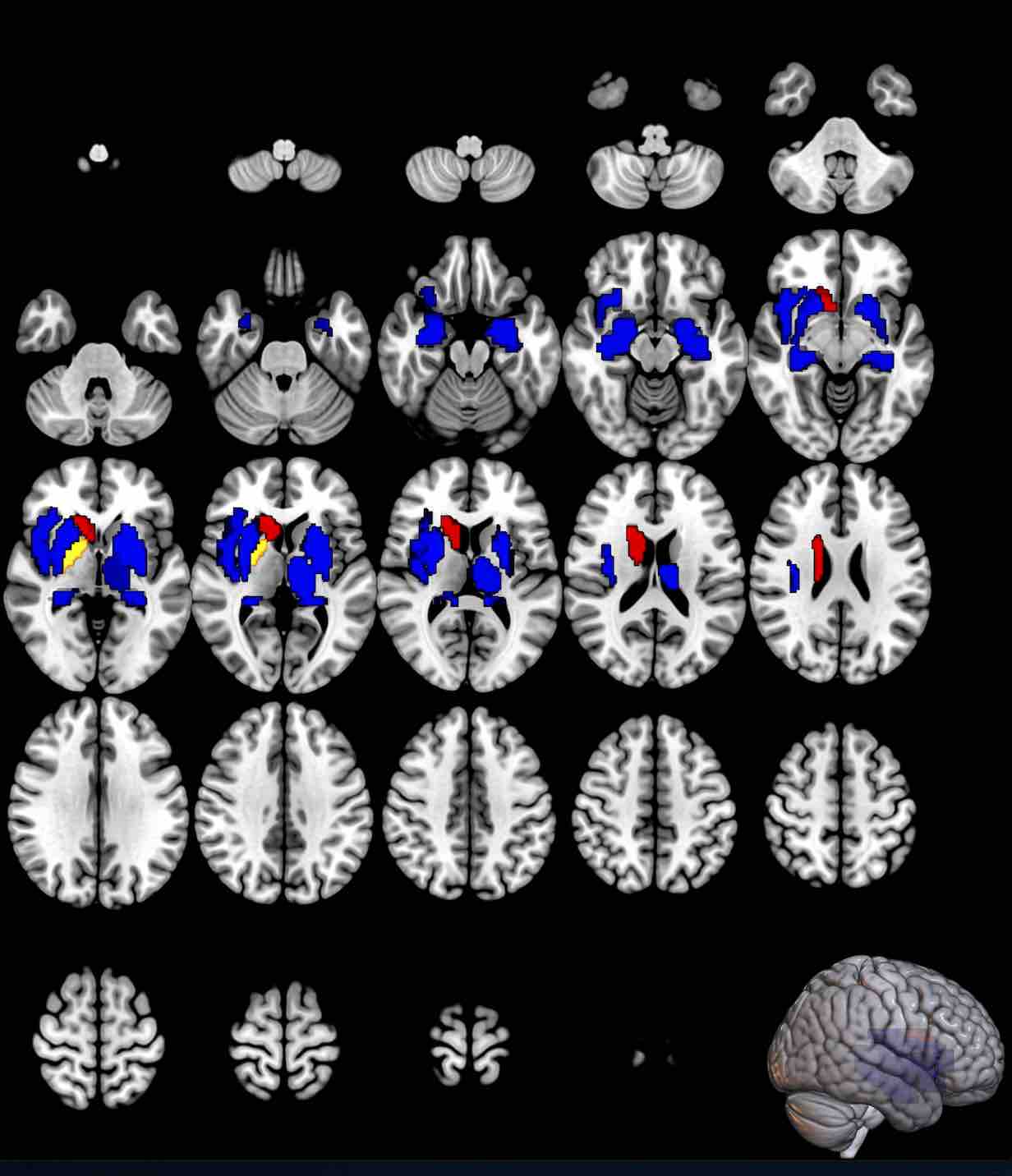}
\caption{}
\end{subfigure}
\begin{subfigure}{0.49\textwidth}
\centering
\includegraphics[width=\textwidth]{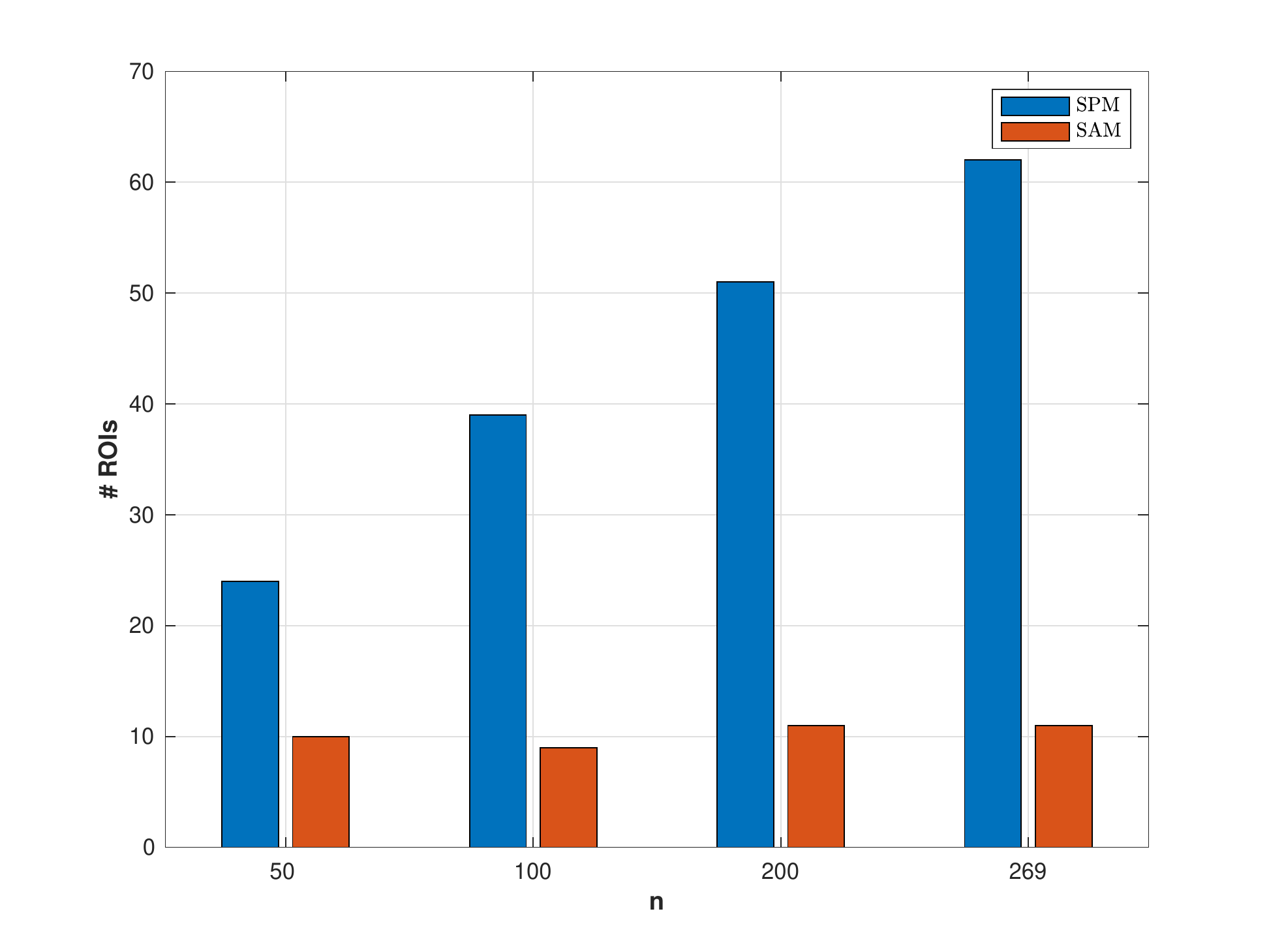}
\includegraphics[width=\textwidth]{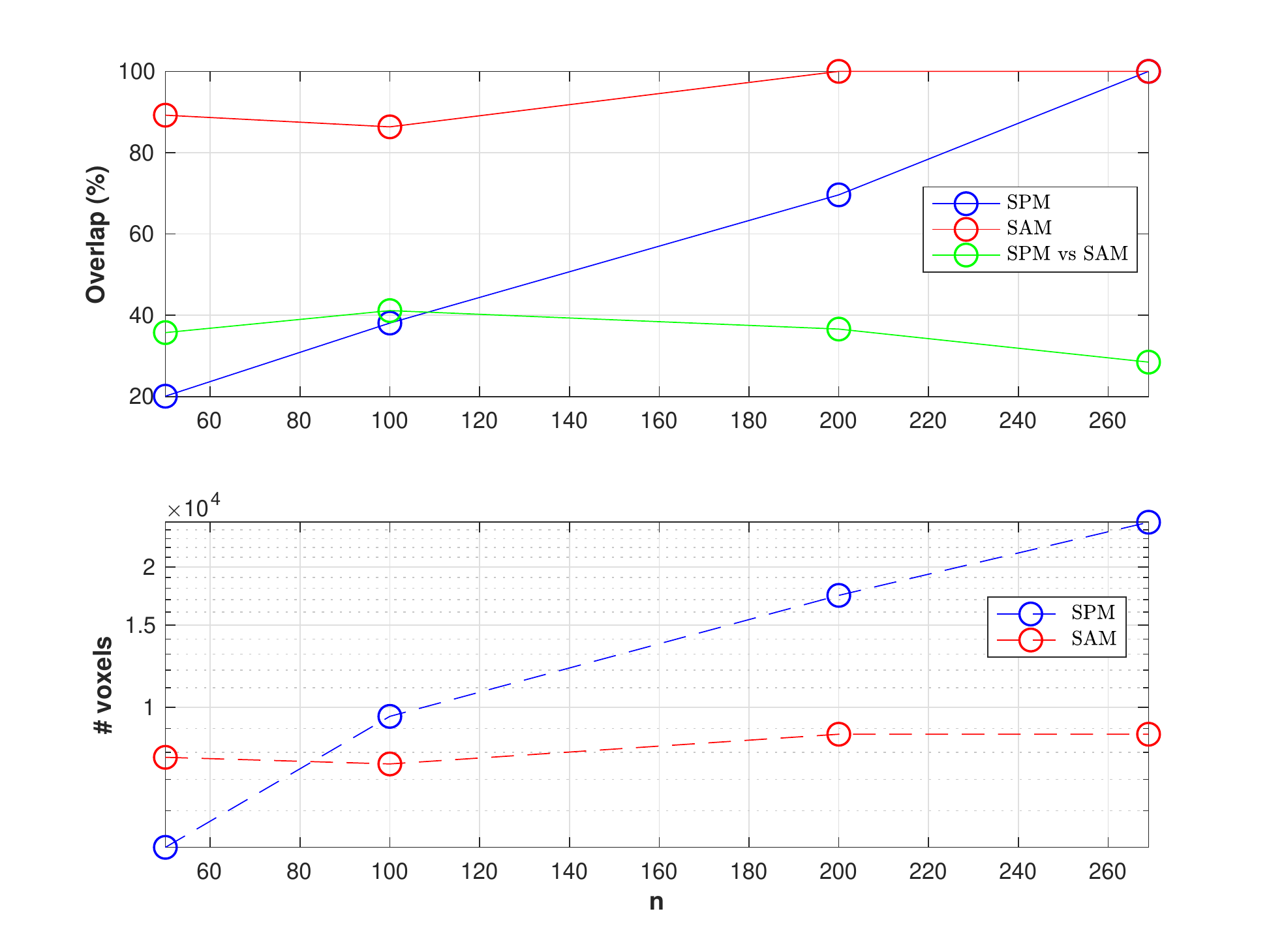}
\caption{}
\end{subfigure}
\caption{(a) SPM (up) and SAM (bottom) using the PPMI dataset for $n=50, 100, 200 and 269$. (b) overlap analysis vs sample size. Observe how the SPM activation map linearly increases with $n$ and is located on more than $60$ standardized regions with the whole dataset. Typical effects, such as PVE, in this kind of low-resolution image modality results in rejecting the null-hypothesis although FWE corrected p-values were considered in the inference test. On the contrary, SLT is less specific but more stable in the rejection of the null-hypothesis. In addition, the ROIs obtained are overlapped more than $80\%$, using a wide range of small sample sizes.}
\label{fig:SAM4}
\end{figure*}

Compared with the subtle effects in the ADNI dataset, the magnitude of the effect in this study is relatively large. Thus, maps of significance derived from both approaches should be similar each other in the specific regions. However, this image modality has associated important challenges such as low resolution empowering partial volume effects (PVE) \cite{Zaidi06} and lack of structural information in the images to perform an accurate spatial normalization and co-registration, \cite{Illan12}. These issues could reveal the limitations of voxel-wise approaches using sharp null hypothesis tests, which may find small effects that are practically unimportant.  All these questions are found in figure \ref{fig:SAM4} where we show how SAM are stable several sample sizes and included in the regions detected by the SPM approach. Moreover, we see how the number of voxels in the classical approach is dramatically rising with increasing $n$, due to the fact that large studies are more likely to find a significant difference for a persistent trivial effect that is not really meaningfully different from the null \cite{Ioannidis05}.

\subsection{An fMRI study: the SPM database}\label{sec:exp3}

Data used in the preparation of this article was obtained from the SPM database related to an epoch auditory fMRI activation data \footnote{www.fil.ion.ucl.ac.uk/spm/data/auditory/}. This database is one of several databases included in the SPM site \footnote{www.fil.ion.ucl.ac.uk/spm/} for personal education and evaluation purposes, and shows the ability of the SPM methodology for detecting auditory stimulation maps. Specifically, the experiment associated with the data was conducted by the FIL methods group and was designed for exploring equipment and techniques related to fMRI. 

The database consists of BOLD/EPI images obtained from a single subject. They were acquired on a modified 2T Siemens MAGNETOM Vision system. The number of acquisitions was $96$ and each one consisted of $64$ contiguous slices ($64\times 64\times 64\times$ $3\times 3\times 3$ mm$^3$ voxels). Acquisition took $6.05$s, with the scan to scan repeat time (TR) set arbitrarily to $7$ s. The acquisitions were made in blocks of $6$, giving $16$ $42$ s blocks. The condition associated with each block alternated between rest and auditory stimulation, starting with rest.  Auditory stimulation was bi-syllabic words presented binaurally, at a rate of $60$ per minute. As the SPM site recommends the first few scans are discarded to avoid T1 effects in the initial scans of an fMRI time series. Then, $84$ acquisitions were finally used after discarding the first complete cycle ($12$ scans). The images were preprocessed (realigned, coregistered using a sMRI, normalized and smoothed) for collecting two different conditions, rest and listening. Then, a GLM specification followed by model estimation and a t-test-based inference (FWE p-value = 0.05) resulted in the activation maps for this auditory-evoked potential experiment. 

\subsubsection{Detecting auditory stimulation maps}

In the last sections we have seen the potentiality of the proposed approach for ROI detection in several binary classification paradigms, i.e. diagnosis, given the usefulness of machine learning. Images collected from the aforementioned experiment are used to identify areas performing a specific information processing function, such as the primary auditory cortex. 

\begin{figure*}
\centering
\begin{subfigure}{0.49\textwidth}
\centering
\includegraphics[width=\textwidth]{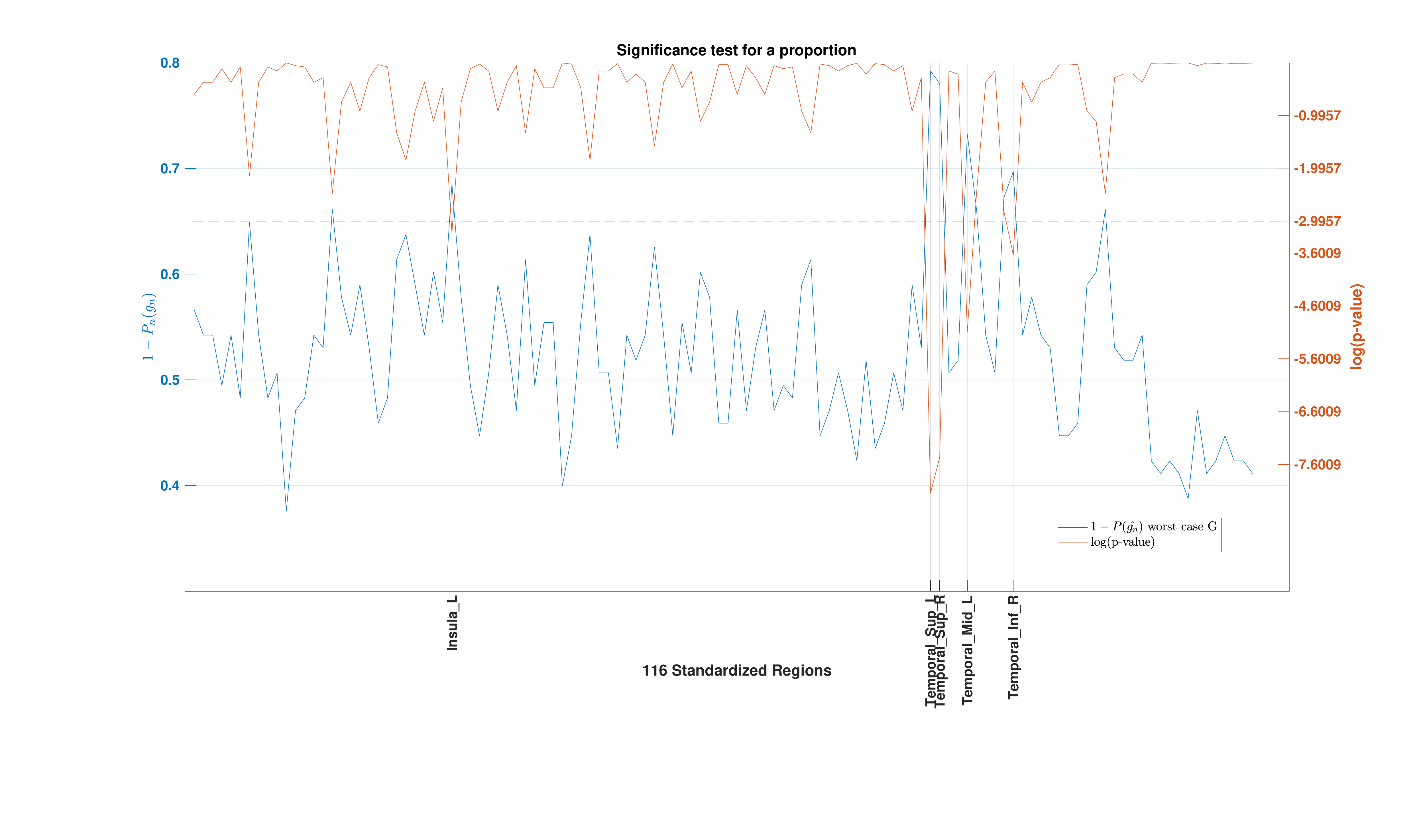}
\includegraphics[width=\textwidth]{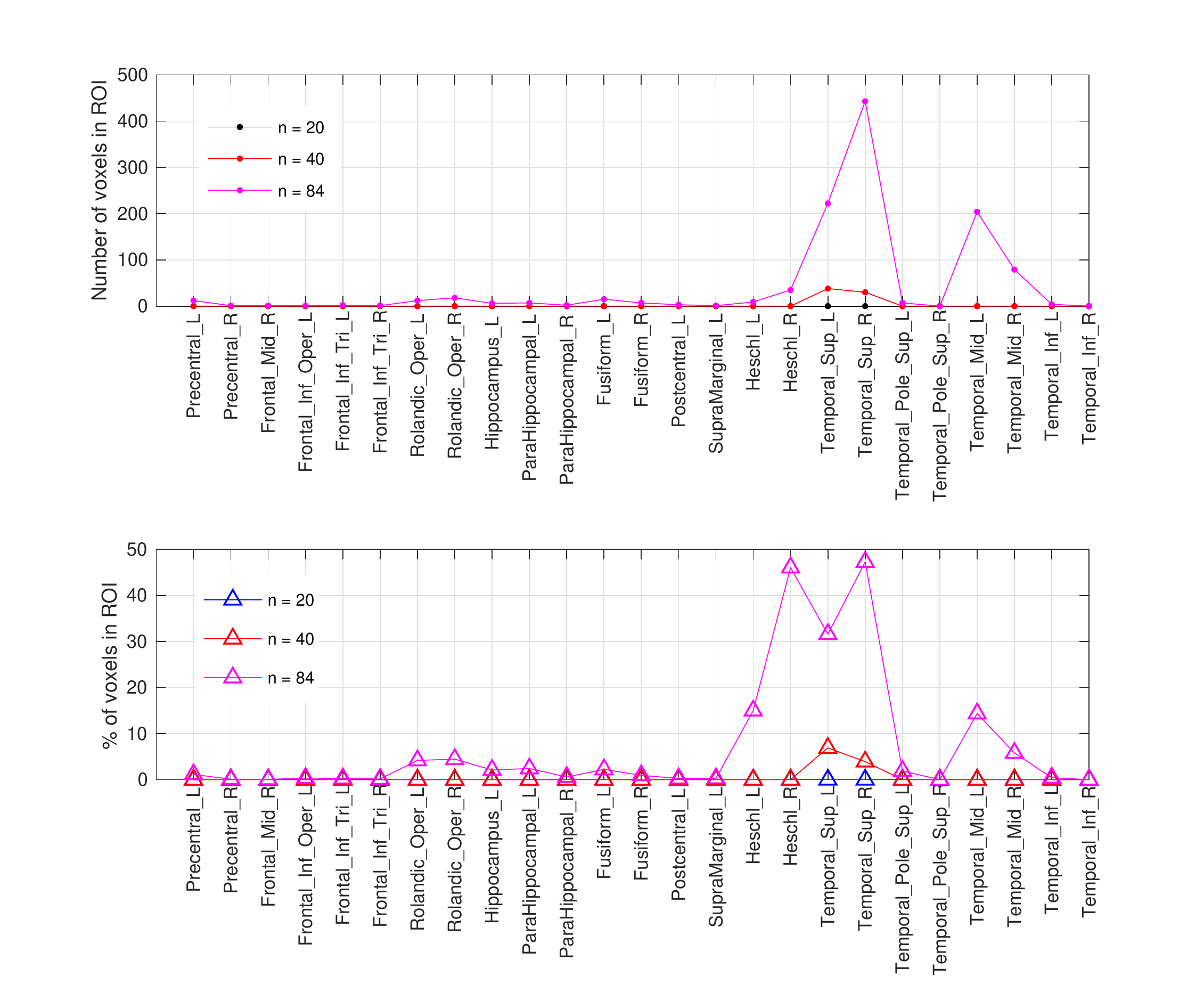}
\includegraphics[width=\textwidth]{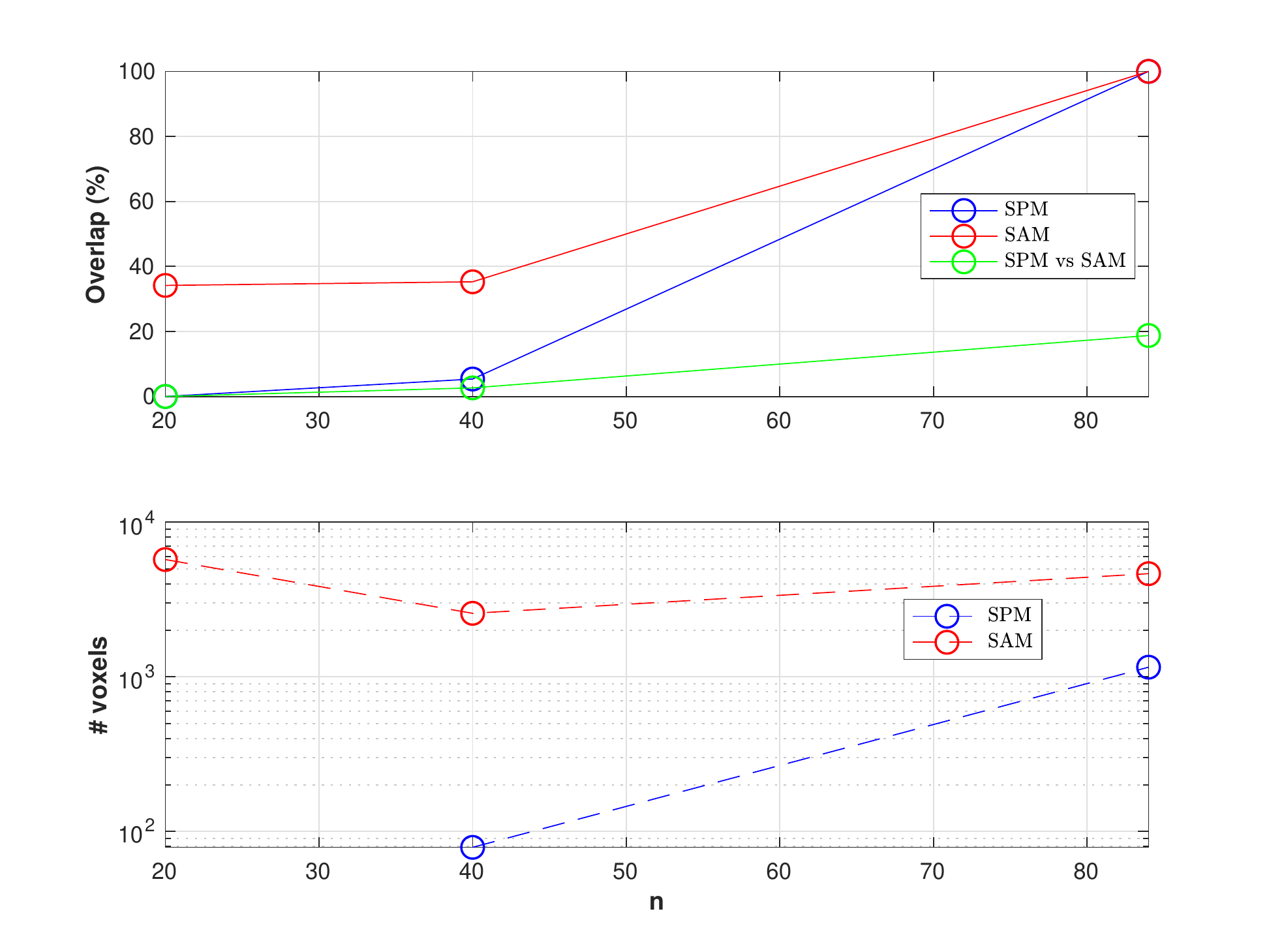}
\caption{}
\end{subfigure}
\begin{subfigure}{0.49\textwidth}
\centering
\includegraphics[width=\textwidth]{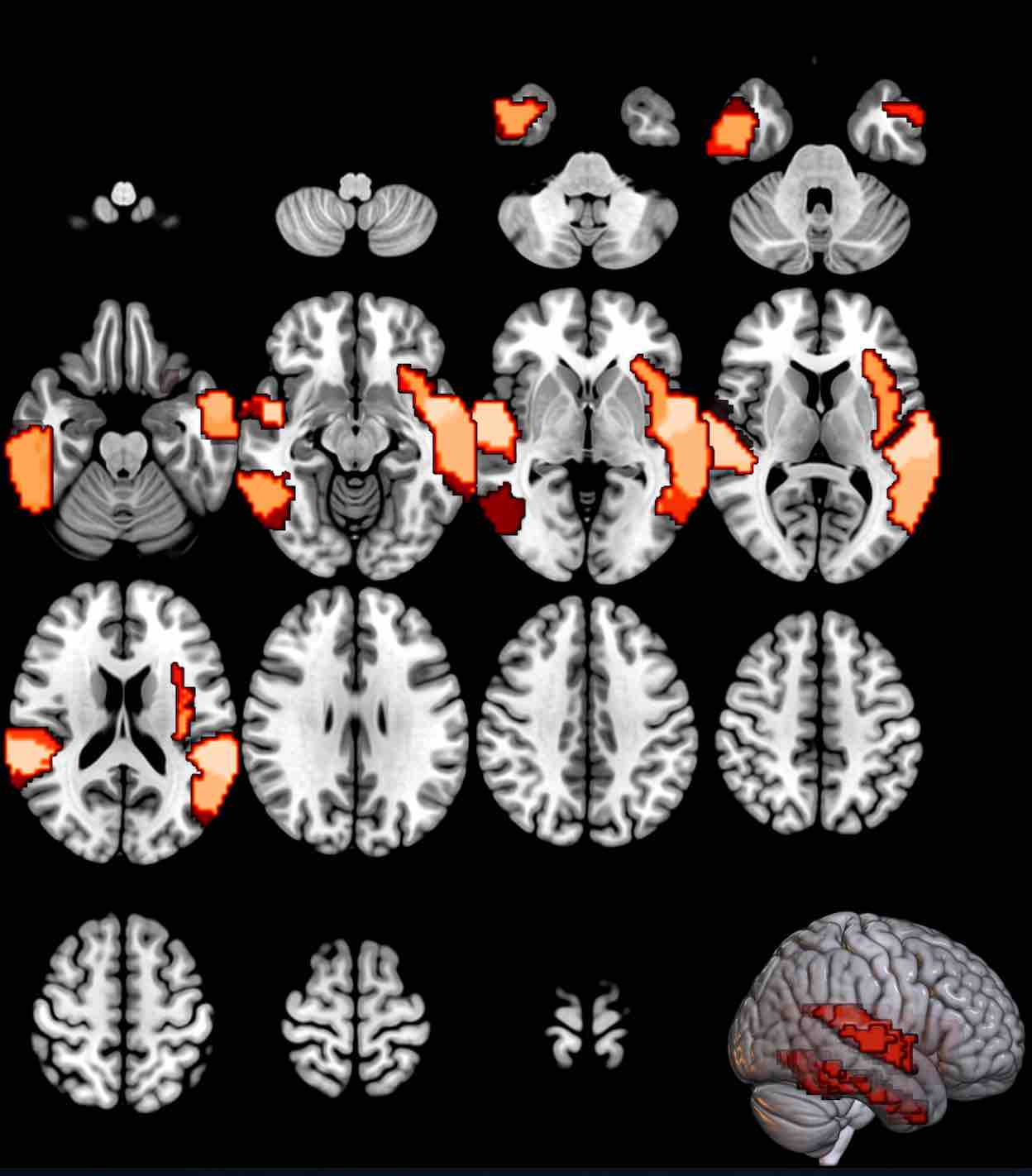}
\includegraphics[width=\textwidth]{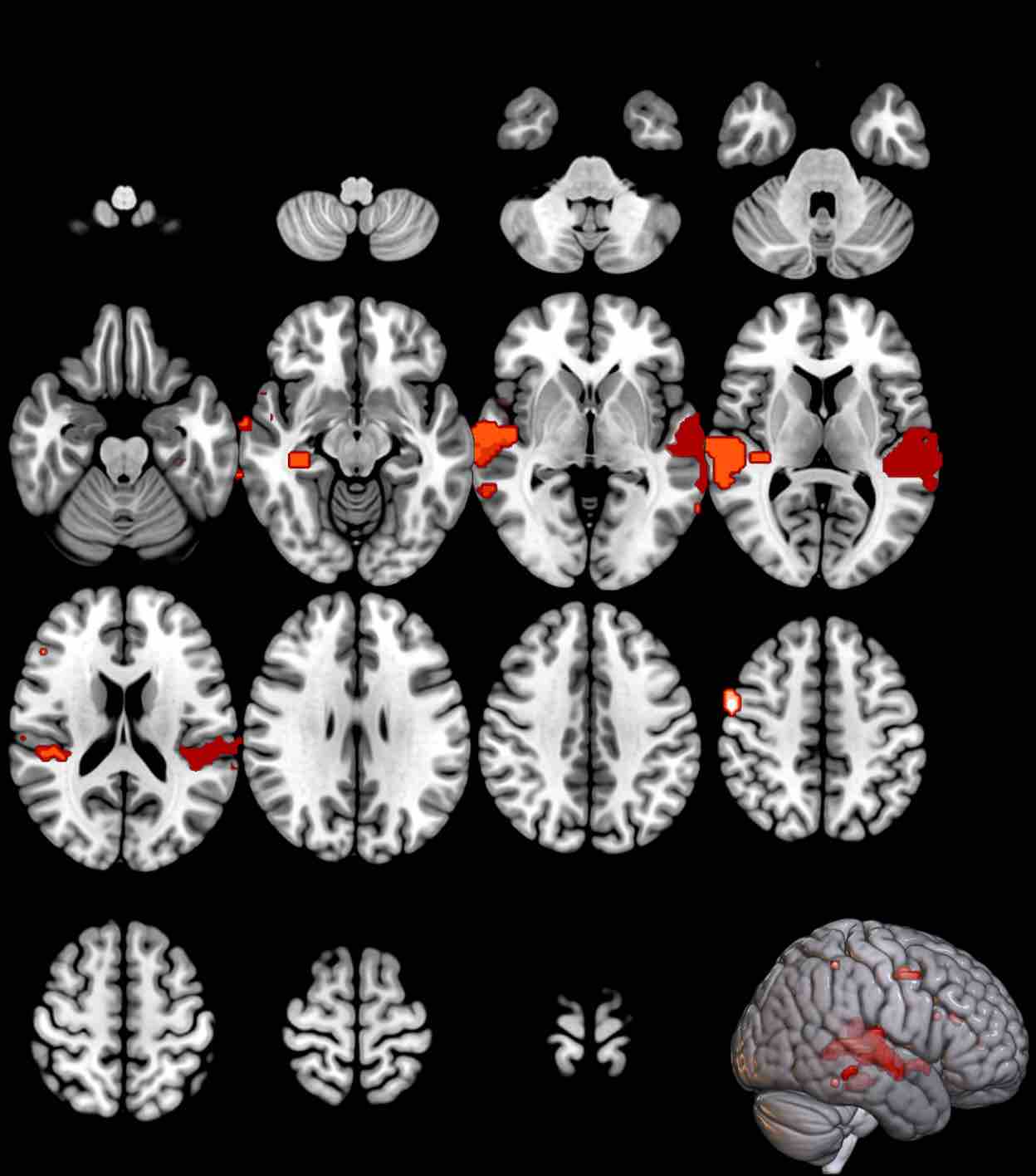}
\caption{}
\end{subfigure}
\caption{(a) Significant tests for a population proportion in the fMRI experiment (up), number of activated voxels in ROIs using SPM (middle) and overlap analysis between SPM and SAM (bottom) (b) Activation maps in the auditory experiment for the whole dataset using the SAM (up) and the SPM (bottom)}
\label{fig:SAM5}
\end{figure*}

The areas identified by the proposed approach are mainly those corresponding with the temporal lobe, as shown in figure \ref{fig:SAM5}. A mosaic and the 3D representation of the activated cortical areas are shown in the same figure \ref{fig:SAM5} (b), together with the activation pattern sought by the SPM methodology. The comparison analysis of both approaches is displayed in figure \ref{fig:SAM5} (a). In the upper figure we see the significance test for a proportion ($n=84$) that was applied to this auditory fMRI experiment. The SAM is mainly located on regions where we found the activation voxels in SPM. In the middle we represent the number of voxels in ROIs (for different sample sizes) and the ratio w.r.t the total number of voxels in that region. Finally, in the bottom we compared both approaches using the overlap-analysis type measures, as described in the last sections. To sum up, we found: i) the same ROIs in both approaches, ii) SPM required sufficiently large sample size to provide significant ROIs, i.e. for $n=20$ no significant areas were sought, and iii) both approaches converge with increasing sample size to the same number of activated voxels.

\section{Discussion}

As shown in the latter section, in general the SAM is a very robust method, in terms of sample size, to find relevant standardized areas, and a stable framework which contains those regions defined as relevant by the SPM, with sufficiently large sample size. The experiments carried out in different experimental frameworks and datasets have demonstrated the ability of this multivariate approach for establishing a novel model-free method for the assessment of significant changes across brain volumes.

The behaviour of the analysed methods depends on the size of effect we are interested in. In the seek of subtle effects, such as the ones found in AD or Autistic patterns, and provided that hypothesis tests cannot separate important, but subtle, and actually trivial effects \cite{Lindquist13}, our SAM focus on standardized ROIs to avoid the presence of false positives in the sought maps. In this sense, SPM is more specific and can detect, within these regions sought by SAM, which substructures are responsible for the discrimination between classes.

On the other hand, when large effects are bound to be found, SAM is a suitable method in their detection since, with a few amounts of samples, it provides similar results than the ones obtained with complete databases, i.e. fMRI and DatSCAN experiments. This is in line with the main idea derived from \cite{Friston12} that when an effect is found in small datasets is more than likely to be extrapolated in large samples. On the contrary, only in small datasets with small – but meaningful – effects that are missed, missing data, sampling bias, etc. we found the absence of replication, i.e. across data collecting sites \cite{Lindquist13}. All these statistical features in the analysis of neuroimaging data are experimentally described in the datasets analysed throughout this paper.

Finally, we have seen the usefulness of the confidence intervals derived for the STL based on concentration inequalities to achieve a confidence framework beyond sharp null-hypothesis testing. Key to this methodology in the field of SLT is that it is based on in-sample estimates (a similar procedure in exploratory analysis using hypothesis testing), unlike the out-sample estimates in CV-procedures, which usually subdivide the (small) datasets for an estimation of the actual error. In this way, an analytical bound depending on sample size ($n$) and number of predictors ($d$) defines a ``worst-case'' operation point. Nevertheless, the experiments showed the application of a systematic hypothesis test for the selection of significant empirical errors which conforms the highlighted regions in the SAM. Only in this case, a model is assumed in the set of accuracies, but it has been demonstrated to be in accordance with the nature of the one-dimensional data and sufficiently accurate for our purposes.

\section{Conclusion}

In this paper we present a data-driven approach, mainly devoted to classification problems with limited sample sizes, to derive statistical model-free (agnostic) mappings.  Although the latter is \emph{not designed for testing competing hypothesis or comparing different models} in neuroimaging, we derive the SAM assuming the existence of classes ($H_1$), at voxel or multi-voxel level. The analysis of the ``worst case' considers the upper bounds of the actual risk, under suitable theoretical conditions (see methods and appendix) and a selection of regions with a highly-corrected empirical risk, according with a test for significance on a population proportion. As a conclusion, the SAM relieved the problem of instability in limited sample sizes, when determining maps of relevance in several neurological conditions, such as AD, PD or auditory tasks, and resulted in a very completive and complementary method with the SPM framework, which is mainly accepted by the neuroimaging community. Moreover, the latter usually employs several strategies for reducing the false positive rates in multiple comparisons, such as the (FWE) corrected p-value maps in inferential statistics null-hypothesis testing, and RFT to tackle with the spatial structure of the maps. However, this approach is found to be very conservative in our experiments and in the extant literature. In this sense, the novel framework based on SLT provides similar activation maps than the ones obtained by the SPM, but defined on ROIs, under a rigorous development in scenarios with a small sample/dimension ratio and large, small and trivial effect sizes, as shown in the experimental part.

\section*{Acknowledgments.} 
This work was partly supported by the MINECO/ FEDER under the RTI2018-098913-B100 project. We would like to thank the reviewers for their thoughtful comments and efforts towards improving our manuscript. J.M. Gorriz would like to thank Dr. Maxim Raginsky for his elegant abstract notation which is borrowed in this paper.

Data collection and sharing for this project was funded by the Alzheimer’s Disease Neuroimaging Initiative (ADNI) (National Institutes of Health Grant U01 AG024904) and DOD ADNI (Department of Defense award number W81- XWH-12-2-0012). ADNI is funded by the National Insti- tute on Aging, the National Institute of Biomedical Imaging and Bioengineering, and through generous contributions from the following: AbbVie, Alzheimer's Association; Alzheimer's Drug Discovery Foundation; Araclon Biotech; BioClinica, Inc.; Biogen; Bristol-Myers Squibb Company; CereSpir, Inc.; Cogstate; Eisai Inc.; Elan Pharmaceuticals, Inc.; Eli Lilly and Company; EuroImmun; F. Hoffmann-La Roche Ltd and its affiliated company Genentech, Inc.; Fujirebio; GE Healthcare; IXICO Ltd.; Janssen Alzheimer Immunotherapy Research $\&$ Development, LLC.; Johnson $\&$ Johnson Pharmaceutical Research $\&$ Development LLC.; Lumosity; Lundbeck; Merck $\&$ Co., Inc.; Meso Scale Diagnostics, LLC.; NeuroRx Research; Neurotrack Technologies; Novartis Pharmaceuticals Corporation; Pfizer Inc.; Piramal Imaging; Servier; Takeda Pharmaceutical Company; and Transition Therapeutics. The Canadian Institutes of Health Research is providing funds to support ADNI clinical sites in Canada. Private sector contributions are facilitated by the Foundation for the National Institutes of Health (www.fnih.org). The grantee organization is the Northern California Institute for Research and Education, and the study is coordinated by the Alzheimer's Therapeutic Research Institute at the University of Southern California. ADNI data are disseminated by the Laboratory for Neuro Imaging at the University of Southern California.

\section*{Supplementary Material}

\begin{figure*}
\centering
\includegraphics[width=1\textwidth]{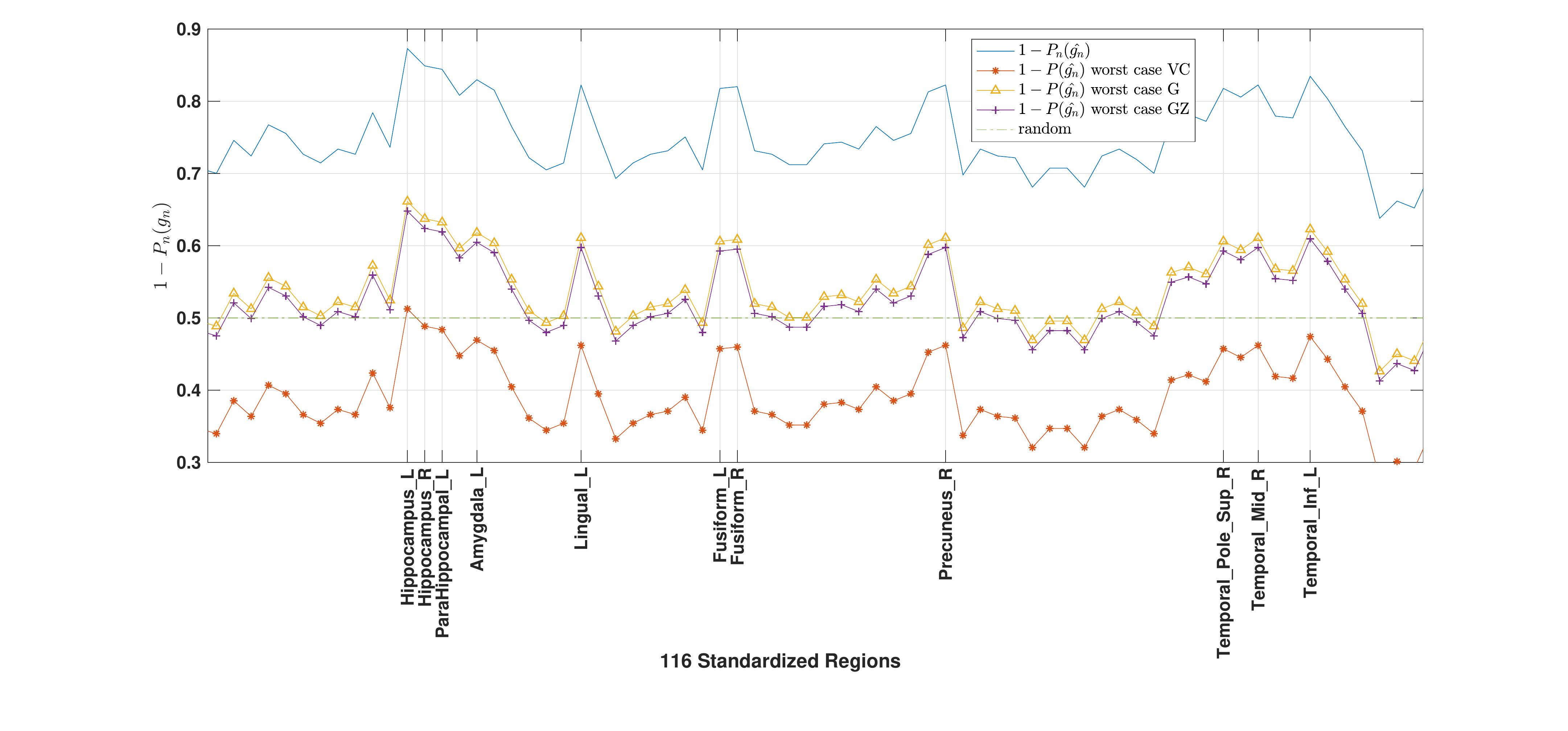}
\includegraphics[width=0.49\textwidth]{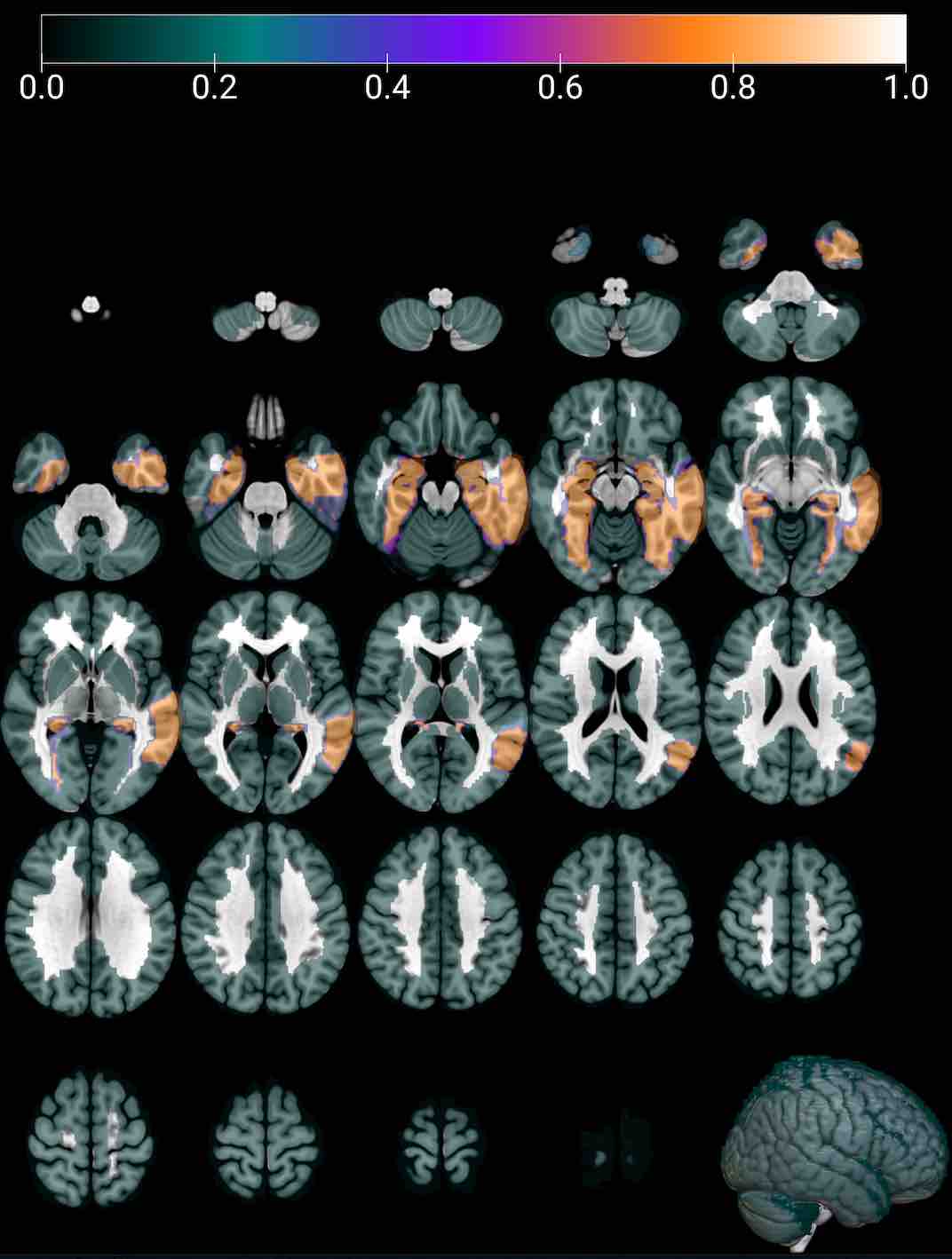}
\includegraphics[width=0.49\textwidth]{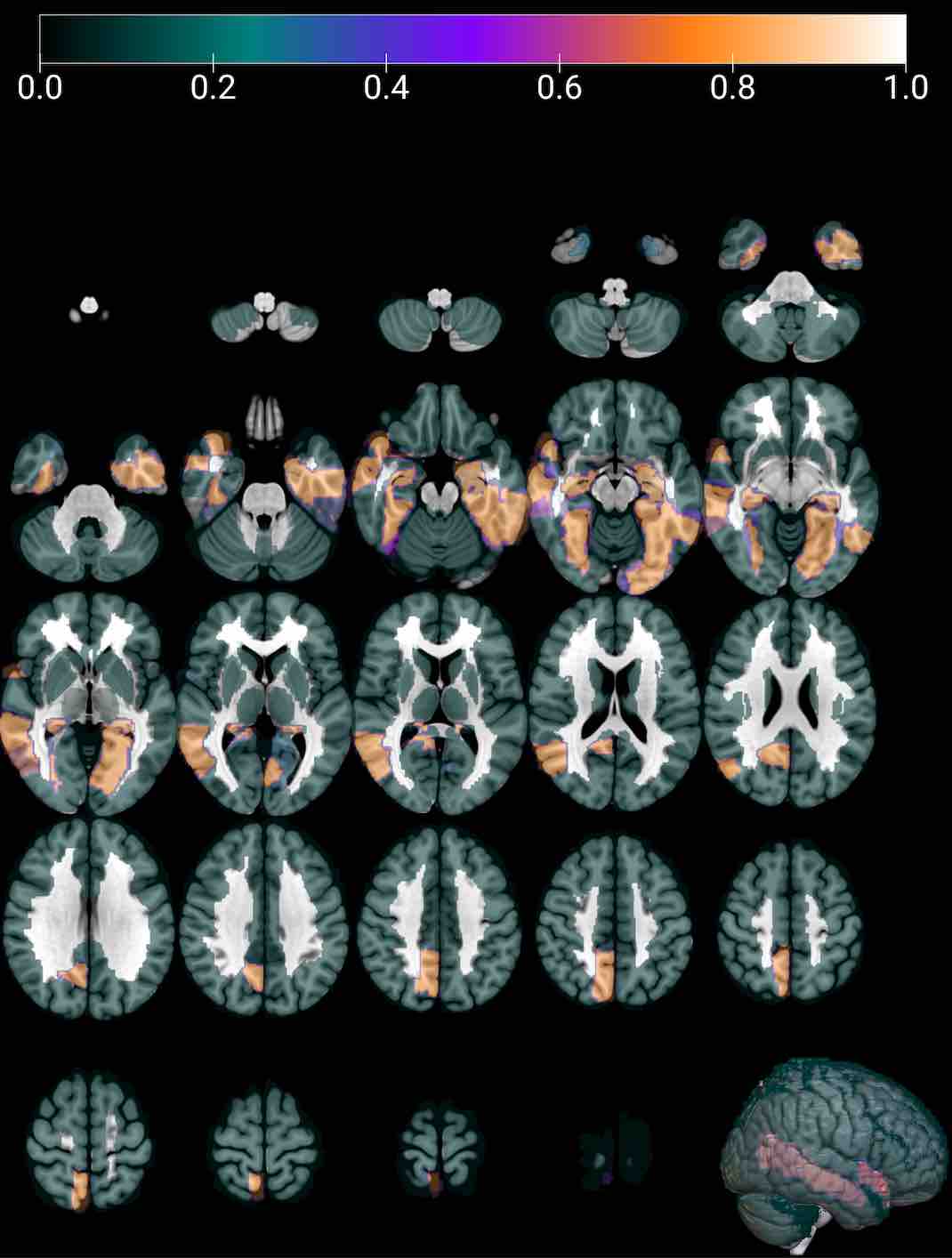}
\caption{The same analysis as the one proposed in figure \ref{fig:class1} but with $d=N_{comp}=8$. Note that the ROIs showing highest accuracy values (bottom on the right) are similar to the ones selected in the latter experiment (bottom on the left) but with an increase in the confidence interval of the approximation. Examples are Hippocampal, Hippocampus, Amygdala and Temporal regions (see section \ref{sec:exp1}).}
\label{fig:class2}
\end{figure*}

A similar analysis was carried out with increasing number of components, i.e. $N_{comp}=2,3,\ldots$, however the upper bounds are increased accordingly as shown in figure \ref{fig:class2}. This highlight the benefits of working in a low dimensional scenario, $d=1$, although the use of new features in the analysis allows us to detect other regions, such as ``Temporal Pole Sup'' and ``Temporal Mid'' regions.

\section*{Appendix}
\subsection*{Upper Bounding the worst case: a summary}
A shown in equation \ref{eq:bound1} the consistency of the ERM algorithm is mainly dependent on the evaluation of the two-sided uniform deviation of the error probabilities in the worst case. An upper bound with probability at least $1-\delta$ for this quantity can be obtained by invoking a result in \cite{McDiarmid89}, since  $\Delta(\mathbf{Z^n})$ has the bounded differences property by $1/n$:
\begin{equation}
\mathbb{P}\left(\Delta(\mathbf{Z^n})\geq\mathbb{E}[\Delta(\mathbf{Z^n})]+t\right)\leq e^{-2nt^2}\equiv \delta;\quad\text{for any}\quad t>0
\end{equation}
This is known as the generalized Hoeffding inequation. Then.,\footnote{Given $x$ a random variable, if $P(x>\epsilon)\leq\eta$ then $P(x<\epsilon)=1-\eta$}, 
\begin{equation}\label{eq:ap1}
\Delta(\mathbf{Z^n})\leq\mathbb{E}[\Delta(\mathbf{Z^n})]+\sqrt{\frac{log(1/\delta)}{2n}}
\end{equation}
with probability $1-\delta$.

Moreover, the expected value of the deviation $\mathbb{E}[\Delta(\mathbf{Z^n})]$ can be absolutely bounded by the so-called Rademacher average \cite{Shalev-Shwartz14} as follows. First, the uniform deviation is bounded by its expected value w.r.t the set of random error functions $g$, using the ``symmetrization'' trick proposed in \cite{Vapnik82} and the convexity property of the norm function:
\begin{equation}\label{eq:ap2}
\Delta_n(\mathbf{Z^n})=\sup\limits_{g\in\mathcal{G}}|P_{n}(g)-P(g)|\leq \mathbb{E}_{g'} [\sup\limits_{g\in\mathcal{G}}|P_{n}(g)-P_n(g')|]
\end{equation}
where $g'$ is randomly selected from $\mathcal{G}$ and $\mathbb{E}_{g'}[P_n(g')]=P(g)$. Taking whole expectations on the both sides we get:
\begin{equation}\label{eq:ap3}
\mathbb{E}[\Delta_n(\mathbf{Z^n})]\leq \mathbb{E} [\sup\limits_{g\in\mathcal{G}}|P_{n}(g)-P_n(g')|]
\end{equation}
By using the triangle inequality and the definition of empirical error we finally obtain:
\begin{equation}\label{eq:ap4}
\mathbb{E} [\sup\limits_{g\in\mathcal{G}}|P_{n}(g)-P_n(g')|]\leq 2 \mathbb{E} [\sup\limits_{g\in\mathcal{G}}|\frac{1}{n}\sum_{i=1}^ng(\mathbf{Z_i})|]
\end{equation}
where the right part of inequality is equally distributed as the Rademacher average $\mathcal{R}(\mathcal{G}(\mathbb{Z^n}))\equiv \mathbb{E}_{\sigma} \sup\limits_{g\in\mathcal{G}} |\frac{1}{n} \sum_{i=1}^n \sigma_i g_i|$, where $\sigma_i$ are independent random variables in $\{\pm 1\}$ with equal probability. Finally, using Massart's finite class lemma \cite{Massart00} we can bound the left part of the latter inequality as:
\begin{equation}\label{eq:ap5}
\mathbb{E}\mathcal{R}(\mathcal{G}(\mathbb{Z^n}))\leq 2\sqrt{\frac{log(N)}{n}}
\end{equation}
Consequently, introducing equations \ref{eq:ap1}, \ref{eq:ap3}, \ref{eq:ap4} and \ref{eq:ap5} in equation \ref{eq:bound1} we finally prove equation \ref{eq:bound2}. 

\subsection*{The partial least squares algorithm}
The PLS algorithm extracts the relevant patterns within ROIs across brains by a regression between the $n\times d$ multivariate data matrix $\mathbf{X}$ and the $n\times 1$ label vector $\mathbf{Y}$. In short, we maximize:
\begin{equation}\label{eq:pls}
\mathbf{\omega}_o=\max_{\omega} \left(cov\left(\mathbf{X}\mathbf{\omega},\mathbf{Y}\right)\right)^2;\quad\text{s.t. }||\omega||=1
\end{equation}
where the score vectors $\mathbf{s}=\mathbf{X}\mathbf{\omega}$ are iteratively extracted and used to deflate the input matrix $\mathbf{X}$ by subtracting their rank-one approximations based on $\mathbf{s}$ \cite{Rosipal06}. The \emph{deflation process} is accomplished by the computation of the vector of loadings $\mathbf{p}$ as a coefficient of regressing $\mathbf{X}$ on $\mathbf{s}$:
\begin{equation}\label{eq:plsloading}
\mathbf{p}=\frac{\mathbf{X}^T \mathbf{s}}{\mathbf{s}^T \mathbf{s}}=\mathbf{X}^T\mathbf{\hat{s}}
\end{equation}
As shown in \cite{Gorriz19} the size of the input data $d$ is crucial to the assessment of the relationship volume data and group membership within the evaluated ROIs, where some statistical properties of the involved processes, such as the stationarity or the ergodicity in the correlation, must be assumed. The PLS-maps derived can be seen as a multivariate two-sample test weighted by the scores of each sample with unknown distribution, except for a normalization term that depends on the pooled standard deviation \cite{Gorriz19}, thus its statistical significance can be assessed in a similar manner of a t-test \cite{Mcintosh96}.

\subsection*{Significance test for a proportion}
Let denote $\hat{\pi}$ the sampling distribution of empirical errors $P_n^i(g_n)$, for $i=1,\ldots,l$, then the null hypothesis test about the population proportion within the confidence interval has the form: 
\begin{equation*}
H_0: \pi=\pi_0 \quad\text{;} \quad H_1:\pi>\pi_0
\end{equation*} 
where $\pi_0$ denotes a particular proportion value between 0 and 1, i.e. 0.5.  The test-statistic in a population proportion is $z=\frac{\hat{\pi-\pi_0}}{\sigma_0}$, where $\sigma_0=\sqrt{(\pi_0(1-\pi_0))/l}$.  For large samples, i.e. for $\pi_0=0.5$ at least $l=20$, if $H_0$ is true, the sampling distribution of the $z$ test statistic is the standard normal distribution.

\bibliographystyle{unsrt}

\begin{thebibliography}{1}

\bibitem{Friston2013}
Friston, K. Sample size and the fallacies of classical inference. NeuroImage 81 (2013) 503–504

\bibitem{Bzdok17}
Bzdok, D. Classical Statistics and Statistical Learning in Imaging Neuroscience. Front. Neurosci., 06 October 2017 | https://doi.org/10.3389/fnins.2017.00543

\bibitem{Vapnik82}
V. Vapnik Estimation dependencies based on Empirical Data.
Springer-Verlach. 1982 ISBN 0-387-90733-5

\bibitem{Kohavi95}
Kohavi, R. A study of CV and bootstrap for accuracy estimation
and model selection. Proc. of the 14th international joint
conference on AI - Vol. 2 pp 1137-1143 (1995)

\bibitem{Gorriz19}
Górriz, J.M. et al. On the computation of distribution-free performance bounds: Application to small sample sizes in neuroimaging. Pattern Recognition 93, 1-13

\bibitem{Gorriz18}
Górriz, et al. A Machine Learning Approach to Reveal the NeuroPhenotypes of Autisms. International journal of neural systems, 1850058

\bibitem{Varoquaux18}
Varoquaux, G. Cross-validation failure: Small sample sizes lead to large error bars. NeuroImage 180 (2018) 68–77.

\bibitem{Friston12}
Friston, K. Ten ironic rules for non-statistical reviewers.  NeuroImage 61 (2012) 1300–1310

\bibitem{Lindquist13}
Lindquist, M.A. et al. Ironing out the statistical wrinkles in "ten ironic rules". Neuroimage. 2013 Nov 1;81:499-502. 

\bibitem{Haussler92}
Haussler, D. Decision theoretic generalizations of the PAC model for neural net and other learning applications. Information and Computation
Volume 100, Issue 1, September 1992, Pages 78-150

\bibitem{Cover65}
Cover, T.M. Geometrical and Statistical properties of systems of
linear inequalities with applications in pattern recognition. IEEE
Transactions on Electronic Computers. EC-14: 326–334 (1965)

\bibitem{Mcintosh96}
McIntosh, A.R. et al. Spatial pattern analysis of functional brain images using partial least squares, Neuroimage 3(3 Pt 1) (1996) 143-157

\bibitem{Rosipal06}
Rosipal, R. et al. Overview and Recent Advances in Partial Least Squares (Springer Berlin, Heidelberg, 2006), pp. 34-51

\bibitem{Rondina15}
Rondina, J.M. SCoRS - a Method Based on Stability for Feature Selection and Mapping in Neuroimaging. IEEE Trans Med Imaging. 2014 Jan; 33(1): 85–98.

\bibitem{Martinez19}
Martinez-Murcia F.J. Studying the Manifold Structure of Alzheimer's Disease: A Deep Learning Approach Using Convolutional Autoencoders. IEEE J Biomed Health Inform. 2019 Jun 17.

\bibitem{DeMartimo08}
De Martino, F. et al. Combining multivariate voxel selection and support vector machines for mapping and classification of fMRI spatial patterns
NeuroImage, 43 (1) (2008), pp. 44-58

\bibitem{Vapnik71}
Vapnik V. et al. On the uniform convergence of relative frequencies of
events to their probabilities. Theory of Probability and Its Applications, 16:264–280, 1971.

\bibitem{Massart00}
Massart, P. Some applications of concentration inequalities to
statistics. Annales de la Faculté des Sciences de Toulouse, 2000.

\bibitem{McDiarmid89}
McDiarmid, C. On the method of bounded differences, Surveys in
Combinatorics 141 (1989), 148–188

\bibitem{Sauer72} 
Sauer, N. On the density of families of sets. Journal of Combinatorial Theory, Series A, 13:145–147, 1972.

\bibitem{Shela72}
Shelah, S. A combinatorial problem: stability and order for models and theories in infinity languages. Pacific Journal of Mathematics, 41:247–261, 1972.

\bibitem{Gomez19}
Gómez-Verdejo, V. Sign-Consistency Based Variable Importance for Machine Learning in Brain Imaging Neuroinformatics October 2019, Volume 17, Issue 4, pp 593–609

\bibitem{Khundrakpam15}
Khundrakpam, B.S. et al. (2015). Prediction of brain maturity based on cortical thickness at different spatial resolutions. NeuroImage, 111, 350–359.

\bibitem{Mouro-Miranda05}
Mouro-Miranda, J. et al. Classifying brain states and determining the discriminating activation patterns: Support vector machine on functional MRI data. NeuroImage, 28, 980–995. (2005).

\bibitem{Friston95}
Friston K. et al. Statistical Parametric Maps in functional imaging: A general linear approach Hum. Brain Mapp. 2:189-210 (1995) 

\bibitem{Friston03} 
Friston, K.J., Harrison, L., Penny, W., 2003. Dynamic causal modelling. Neuroimage 19, 1273–1302.

\bibitem{Reiss15}
Reiss, P.T. Cross-validation and hypothesis testing in neuroimaging: an irenic comment on the exchange between Friston and Lindquist et al. Neuroimage. 2015 August 1; 116: 248–254

\bibitem{Tzourio02}
Tzourio-Mazoyer, N. et al.. Automated anatomical labeling of activations in spm using a macroscopic anatomical parcellation of the MNI MRI single subject brain. Neuroimage 2002; 15: 273-289. DOI

\bibitem{Shalev-Shwartz14}
Shalev-Shwartz, S. et al. Understanding Machine Learning – from Theory to Algorithms. Cambridge University Press. ISBN 9781107057135. 2014

\bibitem{Antos02}
Antós, A. et al. Data-dependent margin-based generalization bounds for classification. Journal of Machine Learning Research 3 (2002) 73–98

\bibitem{Vidyasagar03}
Vidyasagar, M. Learning and Generalisation With Applications to Neural Networks- Springer. ISBN 978-1-84996-867-6 (2003)

\bibitem{Frackowiak04}
Frackowiak et al. Human Brain Function (Second Edition). Chap. 44. Introduction to Random Field Theory. ISBN 978-0-12-264841-0 Academic Press. 867-879, 2004.

\bibitem{NIA18}
Jack, Jr. C.C. NIA-AA Research Framework: Toward a biological definition of Alzheimer’s disease. Alzheimers Dement. 2018 Apr; 14(4): 535–562.

\bibitem{McKhann11}
McKhann, G.M. et al.. The diagnosis of dementia due to Alzheimer’s disease: recommendations from the National Institute on Aging and the Alzheimer’s Assocation Workgroup. Alzheimers Dement. 2011;7:263–9. 

\bibitem{Button13}
Button, K.S. et al. Confidence and precision increase with high statistical power Nature Reviews Neuroscience volume 14, page 585(2013).

\bibitem{Illan12}
Illan, I.A. et al. Automatic assistance to Parkinson's disease diagnosis in DaTSCAN SPECT imaging. Medical Physics. 2012

\bibitem{Zaidi06}
Zaidi, H. et al. Quantitative Analysis in Nuclear Medicine Imaging
Springer Science Business Media, Inc. ISBN-10: 0-387-23854-9

\bibitem{Ioannidis05}
Ioannidis, J.P.A.. Why most published research findings are false. PLoS Med. 2 (8) (e124), 696–701. 2005.

\end{thebibliography}

\end{document}